\documentclass{article} % For LaTeX2e
\usepackage{iclr2027_conference,times}

\usepackage{amsmath,amsfonts,bm}

\def\eqref#1{equation~\ref{#1}}
\def\1{\bm{1}}

\DeclareMathAlphabet{\mathsfit}{\encodingdefault}{\sfdefault}{m}{sl}
\SetMathAlphabet{\mathsfit}{bold}{\encodingdefault}{\sfdefault}{bx}{n}

\usepackage{hyperref}
\usepackage{url}
\usepackage{graphicx}
\usepackage{wrapfig}
\usepackage{caption}
\usepackage{subcaption}
\usepackage{float}
\usepackage{placeins}
\usepackage{booktabs}
\usepackage{capt-of}
\usepackage{tabularx}
\usepackage{array}
\usepackage{cuted}
\usepackage{xcolor}
\usepackage{booktabs} % \toprule, \midrule, \bottomrule
\usepackage{pifont}   % check / cross symbols
\usepackage{amssymb}  % triangle symbol3
\usepackage{longtable}
\usepackage{microtype}
\usepackage{amsmath}

\newcommand{\system}{RePAIR}
\newcommand{\systemfull}{Reinforcement learning with Personalization-Aware Interaction Rewards}
\newcommand{\bench}{PAIR}
\newcommand{\benchfull}{Personalized Application-state Instantiation and Rendering}

\title{Can Agents Work for Everyone? \\ Cross-User Reliability for Mobile GUI Agents in Personalized User Interfaces}

\author{
Yeji Park, Jaeyun Shim, Taesik Gong\\
UNIST \\
\texttt{\{yejipark, tlawodbs1170, taesik.gong\}@unist.ac.kr}
}

\newcommand{\cmark}{\textcolor{green!60!black}{\ding{51}}}
\newcommand{\xmark}{\textcolor{red!80!black}{\ding{55}}}

\iclrfinalcopy % Uncomment for camera-ready version, but NOT for submission.
\begin{document}

\maketitle
% Arxiv
\fancyhead{}                         % 상단 문구 제거
\renewcommand{\headrulewidth}{0pt}    % 상단 가로선 제거

\suppressfloats[t]

\begin{figure}[t]
    \centering
    \includegraphics[width=\linewidth]{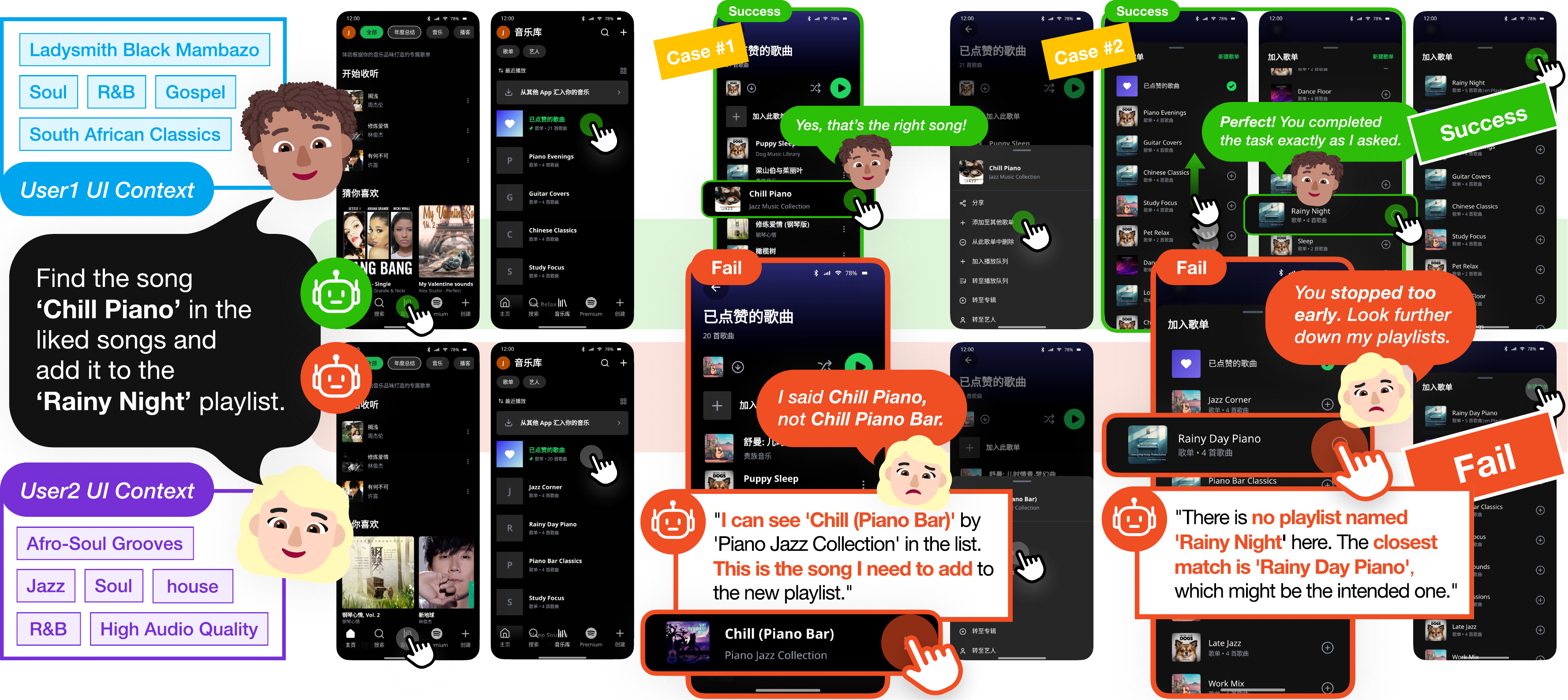}
    \caption{The same request can require navigating different user-conditioned interfaces. We evaluate whether GUI agents remain reliable across users and trace how user-conditioned content disrupts task progress and target selection.}
    \label{fig:main}
\end{figure}

\begin{abstract} 

Mobile GUI agents increasingly operate on interfaces influenced by users' histories and preferences, but their reliability across different users remains underexplored. We introduce \bench{} (\benchfull{}), a pipeline for constructing user-conditioned application states that enables controlled evaluation of the same task across different users. We further introduce \system{} (\systemfull{}), a training approach that learns from cross-user differences in subgoal outcomes to improve reliability across user-conditioned mobile environments. Across six agents, we find substantial variation in task success across users and consistently lower subgoal achievement in user-conditioned UI contexts (6.98 to 15.4 pp). This gap further increases for personal targets drawn from each user's own content (8.77 to 22.0 pp). Failures in these contexts frequently involve selecting another item instead of the intended target, particularly before target exposure. Finally, \system{} improves user-conditioned SAR (+5.87 pp), all-success (+7.50 pp), and overall Task SR (+9.42 pp) over its supervised fine-tuning parent on unseen users, providing initial evidence that explicitly learning from cross-user variation can improve GUI-agent reliability.

\end{abstract}
\section{Introduction}

% 1. State of the world: agents operate in applications shaped by their users.
Recent advances in mobile GUI agents have enabled them to interpret natural-language requests, interpret graphical user interfaces, and perform actions in a human-like manner~\citep{zhou2025mai, qin2025ui, liu2024autoglm, zhao_appagent-pro_2025}. However, GUI agents' action predictions are sensitive to interface presentation, including layout and element positioning ~\citep{yu2026visualattributes,wang2026gui,chen2025less,yang2026focused}. In deployed mobile GUI agents, interface variations naturally arise from users' prior interactions and activity histories. 
% rather than from synthetic perturbations.

% 2. The big BUT: success for one user does not establish reliability for another.
Such user-dependent interface differences can lead to cross-user reliability failures, where an agent that succeeds for one user may fail on the same request for another. 
Figure~\ref{fig:main} shows how user-dependent interfaces can lead an agent to different types of failure, for instance, selecting a visually similar but incorrect target or prematurely concluding that the requested target is unavailable.
Our analysis of users' real mobile interactions from FingerTip 20K~\citep{yang2026fingertip} likewise indicates a 6.41\% (SD 2.38) lower success rate (24.71\% vs. 31.12\%) on labeled target-bearing screen segments containing user-conditioned elements. However, because FingerTip 20K varies both tasks and screen states, it does not support controlled evaluation of the same request across user-conditioned interfaces.

% traces do not align tasks and initial states across users, they cannot demonstrate whether a fixed request remains reliable across user-conditioned interfaces.

% 3. Therefore, we did: persona-based environment generation and controlled analysis.
To address this problem, we introduce \bench{} (\benchfull{}), a pipeline for constructing user-conditioned application states from user profiles. \bench{} extracts application-relevant user preferences, matches them to candidate content for each application based on semantic similarity, selects the highest-ranked content, and renders it in each application's UI.  By varying user-conditioned content while keeping the application structure, \bench{} enables controlled cross-user evaluation of the same task. Additionally, we introduce \system{} (\systemfull{}), a training-based approach to improve cross-user reliability by explicitly learning from differences in agent performance across users. \system{} compares the outcomes of subgoals, goals associated with benchmark-defined UI contexts, across different users' environments under the same task condition and reinforces successful behavior in user-conditioned contexts, where the visible interface content varies across users. By leveraging these cross-user differences as a training signal, \system{} aims to improve consistent task execution across user-conditioned application states.

Using the user-conditioned application states generated by \bench{}, we evaluate whether agents complete the same task consistently across users and where cross-user reliability breaks down. Our evaluation reveals four main findings. First, task success varies substantially across users, with a maximum gap of 9.31 pp between the best- and worst-performing users. 
Second, subgoal achievement is consistently lower in user-conditioned contexts than in user-agnostic contexts, where the visible UI does not depend on the user (+12.0 pp). This gap becomes larger for personal targets where the agent must identify a target drawn from the user’s own content (+14.4 pp). 
Third, in user-conditioned contexts, subgoal failures involve selecting another item instead of the intended target. These errors occur more frequently for personal targets, particularly before target exposure, when additional navigation is required to reach the intended target.
Finally, \system{} improves user-conditioned SAR by 5.87 points, the all-success rate by 7.50 points, and overall Task SR by 9.42 points over its SFT parent on unseen users, suggesting that explicitly training on cross-user differences in user-conditioned subgoals can improve reliability across unseen users.

We summarize our key contributions as follows:
\begin{itemize}
    \item We introduce \bench{}, a user profile-based pipeline for constructing user-conditioned GUI environments that enables controlled evaluation of the same task across different users.
    \item We provide an empirical study of cross-user reliability, showing (i) substantial variation in task success across users, (ii) lower subgoal achievement in user-conditioned contexts, and (iii) more target-selection errors for personal targets, especially before target exposure.
    \item We introduce \system{}, a training approach that leverages cross-user differences in subgoal outcomes to improve reliable task execution across user-conditioned environments.
    % \item We derive design implications for cross-user reliable GUI agents, highlighting the need to account for cross-user variation, adapt interaction to user-conditioned contexts, and strengthen target grounding in personalized interfaces.
\end{itemize}

\section{Related Work}

\begin{table}[t]
\centering
\small
\setlength{\tabcolsep}{2.5pt}
\caption{
Comparison of GUI-agent evaluation settings. The four dimensions are UI Variation, whether UI variations are explicitly evaluated; User Context, whether user preferences or histories are used; User-Conditioned GUI, whether personal records shape the app state; and Cross-User Comparison, whether the same task is evaluated across user-conditioned application states.
}
\label{tab:benchmark_comparison}
\begin{tabular*}{\linewidth}{@{\extracolsep{\fill}}lcccc@{}}
\toprule
Evaluation setting
& UI Variation
& User Context
& \shortstack{User-Conditioned\\GUI}
& \shortstack{Cross-User\\Comparison} \\
\midrule
GUI-Perturbed~\citep{wang2026gui}
& \cmark & \xmark & \xmark & \xmark \\
GUI-Robust~\citep{yang2026guirobust}
& \cmark & \xmark & \xmark & \xmark \\
D-GARA~\citep{chen2026d}
& \cmark & \xmark & \xmark & \xmark \\
\midrule
PSPA-Bench~\citep{nie2026pspa}
& \xmark & \cmark & \xmark & \xmark \\
Persona2Web~\citep{kim2026personaweb}
& \xmark & \cmark & \xmark & \xmark \\
KnowU-Bench~\citep{chen2026knowu}
& \xmark & \cmark & \xmark & \xmark \\
FingerTip 20K~\citep{yang2026fingertip}
& \xmark & \cmark & \cmark & \xmark \\
iOSWorld~\citep{jang2026iosworld}
& \xmark & \cmark & \cmark & \xmark \\
\midrule
\bench{} (Ours)
& \cmark & \cmark & \cmark & \cmark \\
\bottomrule
\end{tabular*}
\end{table}

\subsection{GUI Agent Robustness under Interface Variation}

GUI agents have evolved from grounding instructions in structured GUI representations to perceiving and acting on rendered interfaces~\citep{deng2023mind2web,cheng2024seeclick,wu2025guiactor}. Recent systems integrate visual grounding, reasoning, and action prediction in a unified interaction loop~\citep{zheng2024seeact,qin2025ui}. This shift highlights the need for robustness to variations in interfaces. Visual attributes affect web-agent selections~\citep{yu2026visualattributes}, while zoom perturbations reduce grounding accuracy~\citep{wang2026gui}. Prior methods address GUI complexity through context simplification~\citep{chen2025less} and adaptive visual focus~\citep{ge2025iris}. Meanwhile, benchmarks evaluate robustness using static anomaly datasets~\citep{yang2026guirobust} and dynamically injected disruptions~\citep{chen2026d}. Together, these studies show that reliable GUI interaction requires robustness to dynamic conditions, not just target recognition on fixed screens.

Personalization, however, creates persistent, user-conditioned variation beyond temporary or synthetic perturbations. HCI research shows that usage history and personalized content shape interface presentation~\citep{findlater2004comparison,eslami2015always}, while user-specific records shape the content and context of modern applications~\citep{jang2026iosworld}. Existing robustness studies nevertheless focus on synthetic or transient changes, leaving reliability across persistent, personalized application states underexplored.

% \subsection{Personalization in Mobile GUI Agents}
% Recent personalization work uses preferences, histories, and evolving user context to infer intent or provide proactive assistance~\citep{wang2026meagent,lyu2026personalalign,chai2026pira}. In contrast, we study whether agents can reliably execute a specified action when the interface itself is shaped by the user. 

% Existing evaluations fall into two groups. Robustness benchmarks test visual perturbations~\citep{wang2026gui}, recorded anomalies~\citep{yang2026guirobust}, and live disruptions~\citep{chen2026d}. Personalization benchmarks use preferences, histories, or personal records~\citep{nie2026pspa,kim2026personaweb,yang2026fingertip,jang2026iosworld,chen2026knowu}. As Table~\ref{tab:benchmark_comparison} shows, the compared benchmarks do not directly evaluate agent performance on the same task across user-conditioned application states. We address this gap through controlled cross-user evaluation.

\subsection{Personalization in Mobile GUI Agents}

Existing personalization methods for mobile GUI agents use user preferences, routines, and histories to infer intended actions or provide proactive assistance~\citep{wang2026meagent,lyu2026personalalign}. In contrast, we study whether an agent can reliably execute a fixed instruction when the mobile interface itself varies with user-specific content and state. Prior mobile evaluations examine either robustness to runtime interface disruptions~\citep{chen2026d} or personalized interaction based on user context~\citep{nie2026pspa,yang2026fingertip,jang2026iosworld,chen2026knowu}. However, they do not directly test whether the same instruction and target remain reliable across different user-conditioned mobile application states. We address this gap through controlled cross-user evaluation. As summarized in Table~\ref{tab:benchmark_comparison}.

\section{Motivation: Real-World Mobile GUI Agents Encounter User-Conditioned Interfaces}
\label{sec:motivation}
\label{sec:real_world_observation}

\begin{figure}[t]
    \centering
    \captionsetup{font=normalsize,labelfont=normal}
    \captionsetup[subfigure]{
        font=small,
        labelfont=normal,
        justification=centering,
        singlelinecheck=true,
        skip=2pt
    }

    \begin{subfigure}[t]{0.32\linewidth}
        \centering
        \includegraphics[width=\linewidth]{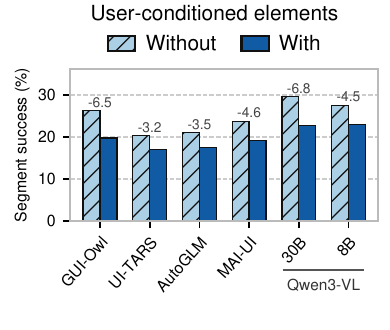}
        \caption{Success with/without user-conditioned elements.}
        \label{fig:ft_segment_success}
    \end{subfigure}\hfill%
    \begin{subfigure}[t]{0.32\linewidth}
        \centering
        \includegraphics[width=\linewidth]{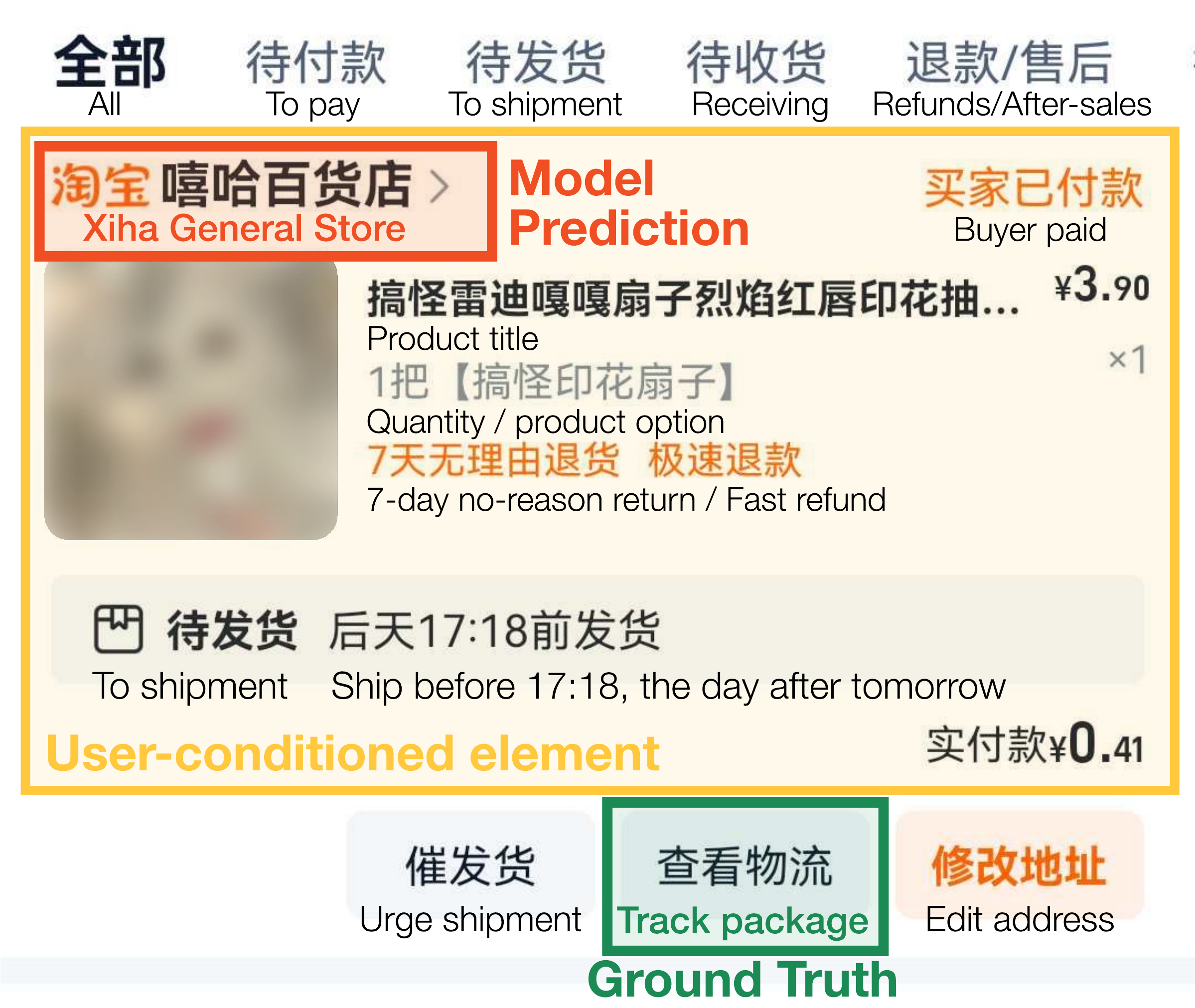}
        \caption{Mistaking user-conditioned order content for the task target.}
        \label{fig:ft_case_douyin}
    \end{subfigure}\hfill%
    \begin{subfigure}[t]{0.32\linewidth}
        \centering
        \includegraphics[width=\linewidth]{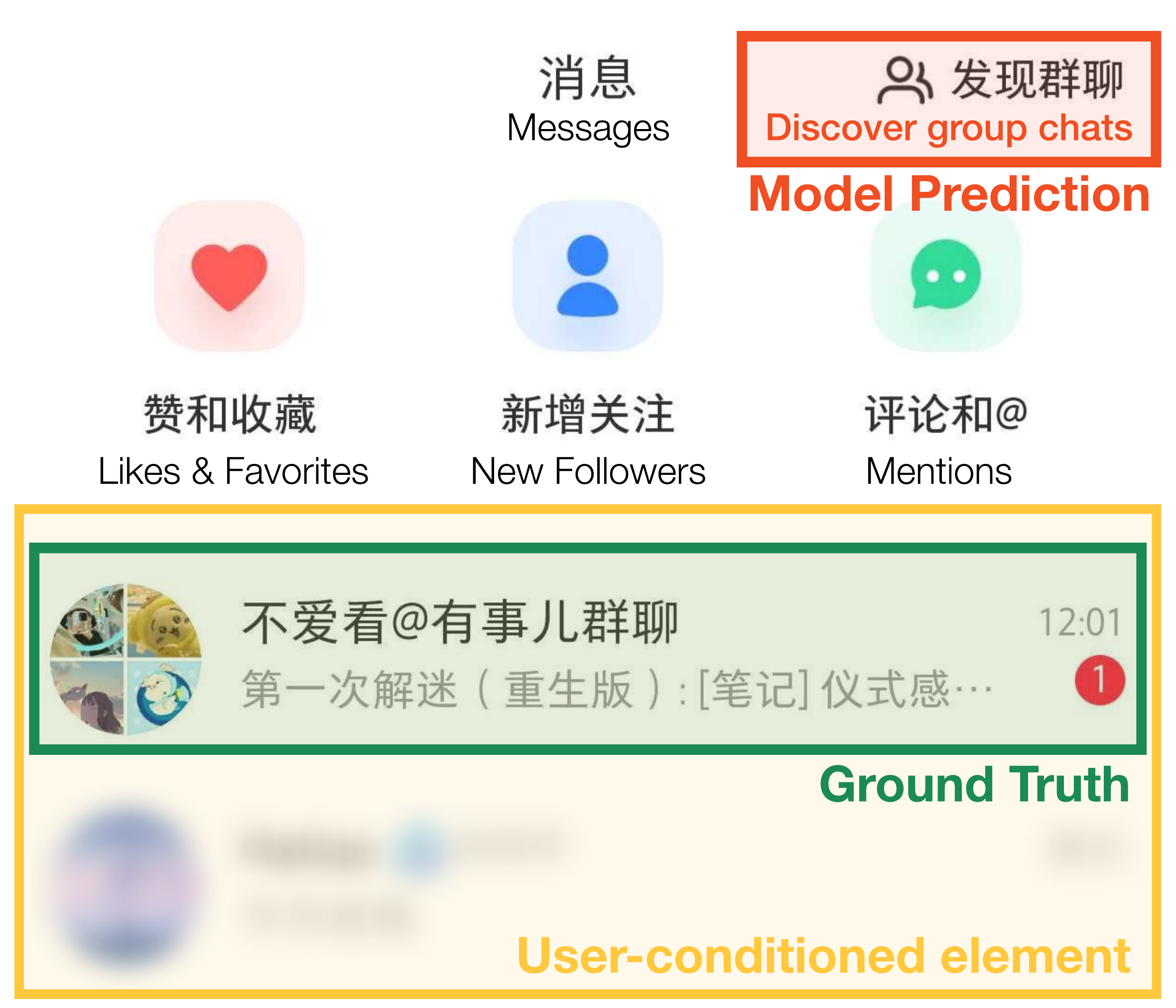}
        \caption{Overlooking the visible user-conditioned unread chat.}
        \label{fig:ft_case_xhs}
    \end{subfigure}
    \vspace{-0.5em}
    % \vspace{-0.4em}
    \caption{GUI-agent behavior on FingerTip 20K. (a) Success by the presence of user-conditioned content. (b,c) Example failures: the agent (b) selects a user-specific shop name instead of Track package and (c) overlooks a user-specific unread chat in favor of Discover group chats. }
    % Red boxes indicate agent actions, green boxes indicate ground-truth actions, and yellow boxes indicate user-conditioned content.}
    
    \label{fig:ft_observation}
\end{figure}

To investigate how user-conditioned content influences mobile GUI agents in everyday interfaces, we evaluate six widely used open-source agent models on Fingertip 20K~\citep{yang2026fingertip}, which captures screenshots and action sequences from users' real mobile interactions. We analyze 103,199 screenshot–action steps spanning 12,016 episodes from 102 apps, collected from 83 users. We group consecutive interactions on the same screen into 60,383 screen segments, and success requires every scored action in a segment to be correct. Across all evaluated agents, segments containing user-conditioned elements, labels unique to a given user within each application, show a mean decrease in success of 4.88 pp (SD 1.50) relative to those without them (Figure~\ref{fig:ft_segment_success}). 

Figures~\ref{fig:ft_case_douyin} and ~\ref{fig:ft_case_xhs} illustrate two types of errors influenced by user-conditioned elements. In the first case, the user asks to check the delivery status of the first order. The user intends to select the order's \textit{Track package} button, but the agent selects the shop name instead, incorrectly assuming it leads to the order details, where delivery information can be found. 
% through the shared \textit{Share link} action, but instead selects a visible user-specific suggested contact. 
In the second case, the user asks to view new messages in a group chat. The intended action is to open the visible group conversation with an unread badge, but the agent selects the shared \textit{Discover group chats} entry instead, incorrectly assuming that it provides access to new messages in the user's group chats.
These examples highlight the need to interpret user-conditioned content in relation to the instruction and identify the appropriate target in the current application state. See more details in Appendix~\ref{app:fingertip_protocol}.
% These examples show that user-conditioned content can redirect agent behavior both toward and away from user-specific items
% \ts{can we visualize user specific items as well in the fig? then it would be more clear - edited}.

These observations suggest that user-conditioned content can interfere with reliable action selection. However, because FingerTip varies tasks and screen states together, it cannot determine whether the same request is completed consistently across different users’ application states. We refer to this consistency as cross-user reliability. We therefore suggest \bench{} to control for the instruction and target while varying user-conditioned application states and study the following questions:

\begin{itemize}
    \item \textbf{RQ1:} How robustly do mobile GUI agents complete the same request across user-conditioned interface states? 
    \item \textbf{RQ2:} Where does task progress break down in user-conditioned interfaces, and how do failure patterns vary with the semantic alignment between the target and the interface?
\end{itemize}
\section{\bench{}: Personalized Application Environment Generation}
\label{sec:environment_generation}

We introduce \bench{} (\benchfull{}), a pipeline that turns user profiles into personalized, executable GUI environments. \bench{} provides a controlled cross-user evaluation environment by pairing user-conditioned application states with matched task templates and success criteria. This section describes the environment generation pipeline, task design, and evaluation protocol. Figure~\ref{fig:environment_generation} illustrates the pipeline overview. 

\subsection{Cross-User Reliability Formulation}
\label{sec:problem_formulation}

Our goal is to assess cross-user reliability, defined as the ability of a mobile GUI agent to execute the same request consistently across different users' interfaces, and to identify where task progress breaks down during execution. Given an instruction $q$ and user $p$'s application state, the agent observes a screenshot $o_t$ and selects an action $a_t$ at each interaction step $t$:
\begin{equation}
a_t = \pi(q,h_t), \qquad s_t=(o_t,a_t),
\label{eq:agent_interaction}
\end{equation}
where $\pi$ denotes the GUI agent, $h_t=(o_0,a_0,\ldots,o_t)$ is the interaction history available to the agent, and $s_t$ denotes the interaction step. Each action updates the application state, and repeating this process yields the interaction trajectory $\tau_p=(s_0,s_1,\ldots,s_{T-1})$.

For each task, let $\mathbf{g}(q)=(g_1,\ldots,g_K)$ denote the ordered subgoals along a reference execution path. Each subgoal $g_k$ specifies an objective associated with a predefined UI context in which that objective can be completed, which we call its \emph{subgoal context}. For user $p$ and subgoal $g_k$, let $I_{p,k}$ denote the set of step indices, ordered by time, that occur in the corresponding subgoal context. We define the context-specific interaction sequence as $\tau_{p,k}=(s_t)_{t\in I_{p,k}}$. Because an agent may revisit the same context multiple times, $I_{p,k}$ may contain non-consecutive indices, and $\tau_{p,k}$ therefore aggregates interactions across repeated visits to the same subgoal context.

In our setting, the visible interface state varies across users while the underlying application structure and interaction logic remain fixed. We distinguish user-conditioned (UC) contexts, where relevant visible content varies across users, from user-agnostic (UA) contexts, where the relevant interface content is shared across users, such as common navigation controls. Accordingly, each subgoal context receives a label $c_{p,k}\in\{\mathrm{UC},\mathrm{UA}\}$: UC when it contains relevant visible user-conditioned elements, and UA otherwise. This labeling groups the corresponding subtrajectories by context type.

\subsection{Environment Generation Pipeline}
\label{sec:generation_pipeline}

% We build on MobileGym~\citep{wu2026mobilegym}}, a browser-hosted simulation platform, preserving the logic and rendering of its simulated applications while varying persona records.
% \ts{this is not the right place of the sentence. should be somewhere in the below paragraph or appendix.}

\begin{figure}[tbp]
    \centering
    \includegraphics[width=1\linewidth]{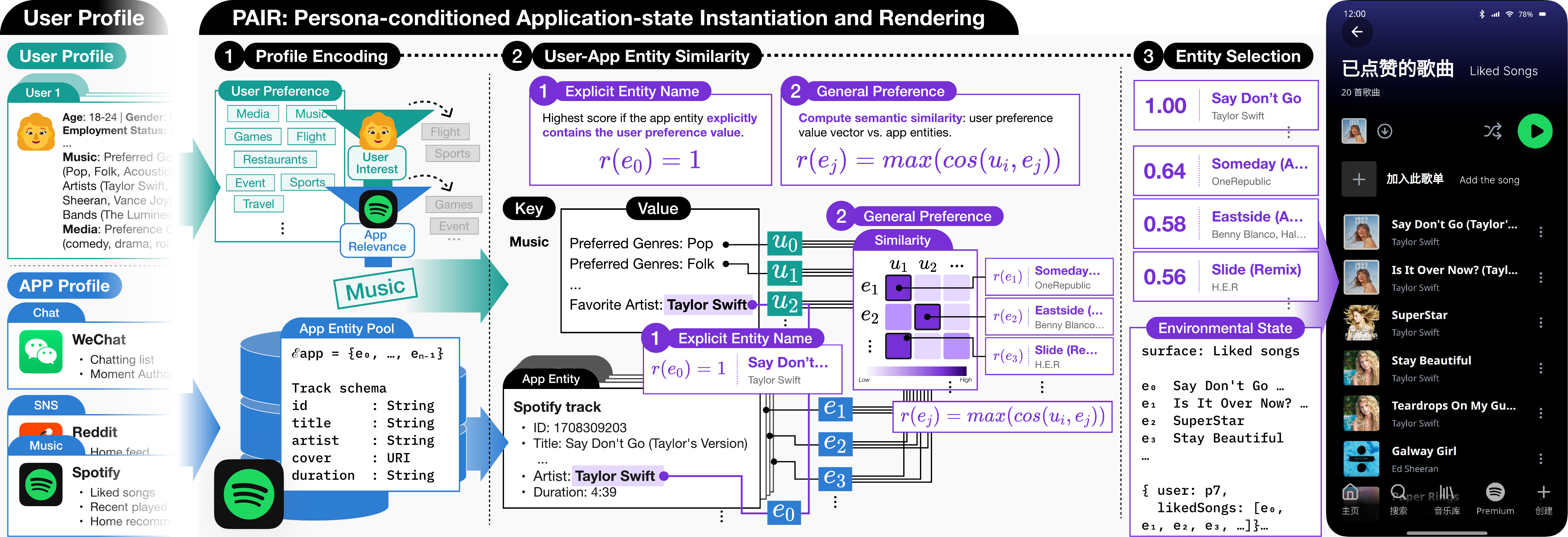}
    \caption{\textbf{Overview of \bench{}.} (i) Given a user and app profile, \bench{} extracts user preferences and filters them according to their relevance to each application. (ii) It then scores candidate application entities using exact matching for explicit entity preferences and semantic similarity for general preferences. (iii) The highest-ranked entities are selected and instantiated in the corresponding application state, which is rendered as a user-conditioned interface.}
    \label{fig:environment_generation}
\end{figure}

% \bench{} takes structured interests and affinities from persona profiles. Additional profiles can instantiate further user-conditioned environments within the supported applications using the same generation process. Each application specifies where user records can appear, including feeds, histories, saved items, chats, and recommendations.
\bench{} constructs user-conditioned application states from interests and attributes derived from user profiles, as illustrated in Figure~\ref{fig:environment_generation}. The framework proceeds in three stages: (i) \textit{profile encoding}, which extracts application-relevant preferences from the user profile and candidate entities, content items that can appear in the application state, from the application profile; (ii) \textit{user-app entity similarity}, which measures their correspondence through exact matching or semantic similarity; and (iii) \textit{entity selection}, which selects the highest-ranked entities to instantiate the application state.
% To support this, \bench{} consists of three key components: (i) extracts application-relevant user preferences and organizes candidate entities from the target application; (ii) measures the correspondence between user preferences and application entities using exact matching or semantic similarity; and (iii) selects the highest-ranked entities to instantiate the persona-conditioned application state.

% \ts{you mention multiple terms here (User Interest Filtering, App Relevance Filtering, etc. you should refer to the fig or define these terms} 
% User Interest Filtering retains active interests; App Relevance Filtering selects those relevant to the application. 
% \subsection{Profile encoding} 
We first apply user interest filtering to retain preferences associated with user $p$'s active interests, followed by app relevance filtering to retain those relevant to application $A$. Let $\mathcal{P}_{p,A}=\{u_i\}$ denote the filtered preference values for user $p$ and application $A$, and let $\mathcal{E}_{A}=\{e_j\}$ denote the candidate entity pool for application $A$.
% \ts{preference entity?}
% entity $e$ available in the application's existing data, we define
% \[z
% s(e,q)=
% \begin{cases}
% 1, & \text{if an explicit entity name directly matches } e,\\
% \cos\!\left(f(e),f(q)\right), & \text{if } q \text{ expresses a general preference},
% \end{cases}
% \]
% where $f(\cdot)$ is a multilingual sentence encoder. Unmatched explicit names contribute no ranking score. The persona relevance of each candidate is then
% \[
% S(e,u,a)
% =
% \max_{q\in Q_{u,a}} s(e,q).
% \]
For each candidate application entity $e_j$, we define its relevance as
\begin{equation}
r(e_j)=
\begin{cases}
1,
& \text{if $e_j$ directly matches any preference value $u_i \in \mathcal{P}_{p,A}$},\\
\displaystyle \max_i \cos\!\left(f(u_i),f(e_j)\right),
& \text{otherwise}.
\end{cases}
\label{eq:entity_relevance}
\end{equation}
where $f(\cdot)$ is a multilingual sentence encoder. 
We rank candidate entities by $r(e_j)$ and use the highest-ranked entities to populate application-specific collections. To instantiate executable environments, we build on MobileGym~\cite{wu2026mobilegym}, injecting the selected entities into user-specific runtime states according to each application's native schema. Structured records are created in their respective schema. The resulting states are rendered by the original application UI.
% and frozen for reproducible evaluation. 
(See Appendix~\ref{app:pair_implementation})

% Candidates with valid scores are ranked by $S(e,u,a)$, and the highest-ranked entities are selected to populate the corresponding persona-conditioned application state. For structured states such as chats, transactions, and meetings, we instead construct persona-specific records directly in the application's native data format.
% \ts{this section should also explain how to generate android environments with the specified scores described above. you may introduce MobileGym here and then how to implement application with system technical terms}

% \ts{you must mention details are in appendix (e.g., prompts, environments, html codes, etc)}

% We distinguish two forms of preference evidence. Some queries specify an explicit entity name, such as a favorite artist or book, whereas others express a general preference, such as a genre, topic, or travel interest. 

\subsection{Task Design}
\label{sec:study_instantiation}

\bench{} constructs task instances by pairing generated user-conditioned states with task templates. Each task template defines the required operation, target UI context, and success criterion, while the target and instruction are varied across conditions.  For each user and template, we instantiate tasks along two axes: \emph{target type} and \emph{instruction type}. For more information, see Appendix~\ref{app:generated_data}, ~\ref{app:task_subgoals}.

\emph{Target type.} 
Targets are either common or personal. A common target is shared across users; its instruction wording is fixed. This enables comparison of the same task across user-conditioned states. In contrast, a personal target is selected from the user-conditioned content generated by \bench{}, testing whether the agent can resolve a target specific to that user.

\emph{Instruction type.}
An explicit instruction refers to the task target by its visible label, whereas an implicit instruction refers to the same target using a contextual cue. For example, an explicit instruction requests, ``Find the post `Looking for interesting flowers for a fashion project!'", whereas an implicit instruction describes the same target as ``a gardening post seeking flower ideas."

\subsection{Evaluation Metrics}

% \paragraph{Metrics.}
\emph{Task success rate (Task SR)} measures end-to-end task completion as $\mathrm{Task\,SR}=E_s/E$, where $E$ is the total number of evaluated tasks and $E_s$ is the number of successfully completed tasks.

\emph{All-success rate} measures cross-user consistency for each template-condition pair, defined as a task template under a fixed target type and instruction type. The all-success rate is defined as $\mathrm{AllSuccess}=M_a/M$, where $M$ denotes the number of template-condition pairs and $M_a$ denotes the number of pairs that achieve all-success; the agent succeeds for every user.
% \emph{All-success rate} measures multi-user task templates in which the agent succeeds for all 10 test users. We define it as $\mathrm{AllSuccess}=M_a/M$, where $M$ is the total number of multi-user evaluation sets and $M_a$ is the number of sets with success across all users. It measures the consistency of task success across users.

\emph{Subgoal achievement rate (SAR)} measures the proportion of these subgoals that are successfully completed. We define it as $\mathrm{SAR}=S_s/S$, where $S$ is the number of evaluated subgoals and $S_s$ is the number successfully completed. SAR enables trajectory-level localization of failures and comparison of subgoal performance between user-conditioned and user-agnostic contexts.

\section{\system{}: Personalization-Aware Interaction Rewards}
\label{sec:method}

To investigate whether rewarding subgoal success in user-conditioned contexts can mitigate performance degradation in mobile GUI agents, we introduce \system{} (\systemfull{}), which uses relative subgoal success across user-conditioned rollouts as the policy optimization signal; see Appendix~\ref{app:reward_implementation} for more detail.

\paragraph{Rollout Group Construction.}
For each combination of task template, target type, and instruction type, we pair training users $p$ and $q$ with compatible subgoal structures. We collect two full-task rollouts from each user's task instance, forming the group $\mathcal G_{p,q}=(\tau_1,\tau_2,\tau_3,\tau_4)$, where $\tau_1$ and $\tau_2$ come from $p$ and $\tau_3$ and $\tau_4$ from $q$. Each rollout starts from the corresponding initial state.

\paragraph{Personalization-Aware Rewards.}
For rollout group $\mathcal G_{p,q}$, let $\mathcal P$ denote its shared set of subgoals in user-conditioned contexts; subgoals in user-agnostic contexts are excluded. For rollout $\tau_j$ and subgoal $k\in\mathcal P$, we assign a binary success reward:
\begin{equation}
r_{jk}
=
\begin{cases}
1, & \text{if }\tau_j\text{ satisfies the criteria for subgoal }k, \\
0, & \text{otherwise}.
\end{cases}
\label{eq:personalization_reward}
\end{equation}
We determine subgoal success from recorded trajectory evidence, including entry-screen observations, visited screens, clicks, and task-judge outcomes.

\paragraph{Group-Relative Advantages.}
For each subgoal, we normalize its rewards across the four rollouts and then average the normalized rewards across user-conditioned subgoals to obtain one rollout-level advantage. Let $\mu_k$ and $\sigma_k$ be the mean and population standard deviation of the rewards for subgoal $k$. For nonempty $\mathcal P$, we compute
\begin{equation}
A_j
=
\frac{1}{|\mathcal P|}
\sum_{k\in\mathcal P:\,\sigma_k>0}
\frac{r_{jk}-\mu_k}{\sigma_k}.
\label{eq:group_relative_advantage}
\end{equation}
For example, suppose $\mathcal P$ contains two subgoals. Across the four rollouts, the first subgoal has reward vector $(1,1,0,0)$ and thus normalized reward vector $(1,1,-1,-1)$. The second has reward vector $(1,1,1,1)$ and thus contributes zero to every rollout. Averaging the two subgoal-level values for each rollout gives $(A_1,A_2,A_3,A_4)=(0.5,0.5,-0.5,-0.5)$. If $\mathcal P$ is empty, we set $A_j=0$.

\paragraph{Policy Optimization.}
We optimize an SFT-initialized policy $\pi_\theta$ with GRPO~\citep{shao2024deepseekmath} using the rollout groups and advantages defined above, while keeping the SFT policy fixed as the reference $\pi_{\mathrm{ref}}$. Let $\pi_{\mathrm{old}}$ denote the rollout-generating policy. For rollout $\tau_j$, let $N_j$ be the number of generated output tokens. At each token position $t\in\{1,\ldots,N_j\}$, $w_{jt}$ denotes the generated token and $h_{jt}$ the instruction and observation--action history available before generating it. We maximize
\begin{equation}
\begin{aligned}
\mathcal J(\theta)
&=
\mathbb E_{\mathcal G_{p,q}\sim\pi_{\mathrm{old}}}
\frac{1}{4}\sum_{j=1}^{4}
\frac{1}{N_j}\sum_{t=1}^{N_j}
\\[2pt]
&\quad
\bigg[
\min\bigg(
\frac{\pi_\theta(w_{jt}\mid h_{jt})}
{\pi_{\mathrm{old}}(w_{jt}\mid h_{jt})}A_j,
\operatorname{clip}\!\bigg(
\frac{\pi_\theta(w_{jt}\mid h_{jt})}
{\pi_{\mathrm{old}}(w_{jt}\mid h_{jt})},
1-\varepsilon,1+\varepsilon
\bigg)A_j
\bigg)
-\beta D_{\mathrm{KL},jt}
\bigg].
\end{aligned}
\label{eq:repair_objective}
\end{equation}
Here $\varepsilon$ sets the clipping range, while $\beta$ controls the KL penalty from the SFT reference. For each sampled token, the penalty is
\begin{equation}
D_{\mathrm{KL},jt}
=
\frac{\pi_{\mathrm{ref}}(w_{jt}\mid h_{jt})}
{\pi_\theta(w_{jt}\mid h_{jt})}
-
\log\frac{\pi_{\mathrm{ref}}(w_{jt}\mid h_{jt})}
{\pi_\theta(w_{jt}\mid h_{jt})}
-1.
\label{eq:kl_penalty}
\end{equation}
% Each rollout is weighted equally, and its advantage applies to all generated response tokens (Appendix~\ref{app:experiment_details}).
Each rollout is weighted equally, and its advantage applies to all response tokens.

\section{Experiments}
\label{sec:experiments}

\begin{table}[t]
\centering
\footnotesize
\caption{Overall performance of base mobile GUI agents and \system{} on \bench{}. Task SR measures task success, all-success measures the rate of tasks completed successfully by all user profiles, and SAR measures subgoal achievement over reached contexts. Parentheses show \system{}'s percentage-point gains over SFT.}
\label{tab:main_results}
\setlength{\tabcolsep}{3pt}
% Approved p0--p9 grid: 30 task templates x 4 conditions x 10 personas.
% Task SR and SAR retain their existing final-goal / entered-context contracts.
% All-success: results/analysis/personalbench_persona_pass_autoglm_greedy_20260920/all_success.csv
% Each target type has 60 multi-user evaluation sets, with ten personas per set.
% All-success pools both target types (120 sets), using integer success counts.
% All-success is PersonaPass^10. In grouping rows parentheses give delta vs SFT (pp).
% UI-TARS Task SR bounds retain six unknown episodes; all-success is exact:
% both sets containing unknown outcomes already contain known failures.
% SAR parentheses are adjudicated-subgoal counts, not changes.
% Panel B retains the base-model row and the matched SFT parent.
% Original evaluation snapshots and final-goal scoring corrections are preserved.
\begin{tabular*}{\linewidth}{@{\extracolsep{\fill}}lccccc@{}}
\toprule
Model
& Params.
& \shortstack{Task SR\\(\%)}
& \shortstack{All-success\\(\%)}
& \shortstack{User-agnostic\\SAR (\%)}
& \shortstack{User-conditioned\\SAR (\%)} \\
\midrule
UI-TARS-1.5~{\scriptsize\citep{qin2025ui}}
& 7B & 48.5 & 22.5 & 79.7 & 65.4 \\

Qwen3-VL-Instruct~{\scriptsize\citep{bai2025qwen3}}
& 30B-A3B & 47.6 & 19.2 & 79.3 & 67.2 \\

Qwen3-VL-Instruct~{\scriptsize\citep{bai2025qwen3}}
& 8B & 47.5 & 22.5 & 74.6 & 67.6 \\

AutoGLM-Phone~{\scriptsize\citep{xu2026mobilerl}}
& 9B & 53.7 & 21.7 & 85.3 & 69.9 \\

MAI-UI~{\scriptsize\citep{zhou2025mai}}
& 8B & 53.4 & 25.8 & 86.2 & 71.0 \\

GUI-Owl-1.5-Think~{\scriptsize\citep{xu2026mobile}}
& 8B & 49.5 & 16.7 & 77.4 & 69.1 \\
\midrule
Qwen3-VL-Instruct
& 8B & 47.5 & 22.5 & 74.6 & 67.6 \\

\quad + SFT
& 8B + 15.3M & 55.1 & 20.0 & 92.3 & 71.9 \\

\quad \textbf{+ SFT + \system{} (ours)}
& 8B + 15.3M
& \textbf{64.5 (+9.4)}
& \textbf{27.5 (+7.5)}
& \textbf{94.3 (+2.0)}
& \textbf{77.7 (+5.9)} \\
\bottomrule
\end{tabular*}
\end{table}

We evaluate cross-user reliability at both the task and trajectory levels using \bench{}. Specifically, we examine (1) how task success and (2) how subgoal achievement are affected across user-conditioned application states, (3) when incorrect-item selections occur relative to target visibility, and (4) whether \system{} improves performance by rewarding subgoal success in user-conditioned contexts.

% We evaluate cross-user reliability, from overall task outcomes to trajectory-level failures using \bench{}. Specifically, we examine (1) how task success varies across persona-conditioned application states, (2) how subgoal achievement changes in user-conditioned contexts, (3) when competing-item errors occur relative to target exposure, and (4) whether RePAIR improves performance by rewarding subgoal success in user-conditioned contexts.

\subsection{Experimental Setup}
\label{sec:experimental_setup}

Using user profiles from PersonaLens~\citep{zhao2025personalens}, we evaluate \bench{} by instantiating ten profiles (p0 to p9) with \bench{} as the test set for our main evaluation. We then instantiate ten additional profiles (p10 to p19) for the \system{} study, using eight for training and two for validation. 
For each user profile, \bench{} instantiates three task templates across ten applications under two target types and two instruction types, yielding $3 \times 10 \times 2 \times 2 = 120$ instantiated tasks per persona.
% For each persona, \bench{} generates task instances from three templates across ten applications, with two target types and two instruction types, yielding $3 \times 10 \times 2 \times 2 = 120$ instances per persona. 
This results in 1,200 test tasks, 960 training tasks, and 240 validation instances. 
% The test personas are disjoint from the training and validation personas, and the frozen test set is used for agent comparison and trajectory analysis.

\paragraph{Models.}
% \ts{first check the model sizes from recent mobile agent papers, and if their sizes are significantly bigger than ours, explain why we focus on relatively smaller backbones  - edited}
We evaluate Qwen3-VL-8B, GUI-Owl-1.5-8B-Think~\citep{xu2026mobile}, MAI-UI-8B~\citep{zhou2025mai}, AutoGLM-Phone-9B-Multilingual~\citep{liu2024autoglm,xu2026mobilerl}, and UI-TARS-1.5-7B~\citep{qin2025ui}, which fall within a deployment-relevant model scale for mobile settings, together with Qwen3-VL-30B-A3B~\citep{bai2025qwen3} as a larger-scale comparison, on the same frozen test set. For the ~\system{}, we first fine-tune Qwen3-VL-8B with LoRA~\citep{hu2021lora} on explicit-instruction demonstrations from the training personas. We then apply \system{} to tasks selected based on the preceding cross-persona analysis. More details are provided in Appendix~\ref{app:experiment_details}.

\subsection{Results}
\label{sec:results}

% \begin{wrapfigure}{l}{0.32\textwidth}
%     \vspace{-\intextsep}
%     \centering
%     \includegraphics[width=\linewidth]{figures/personalbench/all30_ip_failure_compact.pdf}
%     \caption{
%         [].
%     }
%     \label{fig:all30_ip_failure_compact}
%     \vspace{-\intextsep}
% \end{wrapfigure}

To assess overall task completion, cross-user consistency, and subgoal progress, we examine the task success rate (Task SR), the all-success rate, and the subgoal achievement rate (SAR) (see Table~\ref {tab:main_results}). Across the six agents, mean Task SR is 50.0\% (SD 2.82). As we expected, task success decreases when the instruction is implicit rather than explicit (40.7\% vs.\ 59.3\%; $p<.001$). Personal targets also show lower task success than common targets (46.6\% vs.\ 53.5\%). Neither this difference ($p=.0642$) nor the instruction–target interaction ($p=.283$) reached statistical significance.

% Personal targets also show lower task success than common targets (XX.X\% vs.\ XX.X\%), although this difference is not statistically significant ($p=.XXXX$). We observe no significant interaction effect between instruction and target type ($p=.XXXX$).

\begin{figure*}[t]
    \centering

    % =========================
    % Figure 5
    % =========================
    \begin{minipage}[t]{0.32\textwidth}
        \centering

        \includegraphics[width=\linewidth]
        {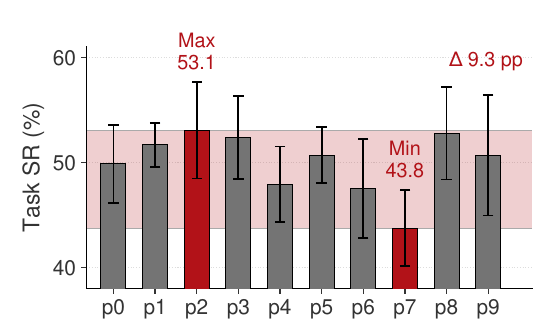}

        \captionof{figure}{
            User-level task success rate(Task SR). Error bars show standard deviation across agents.
            % The min and max differ by 9.3 percentage points.
        }
        \label{fig:acrosspersona}
    \end{minipage}
    \hfill
    % =========================
    % Figure 4
    % =========================
    \begin{minipage}[t]{0.66\textwidth}
        \centering

        \begin{subfigure}[t]{0.48\linewidth}
            \centering
            \includegraphics[width=\linewidth]
            {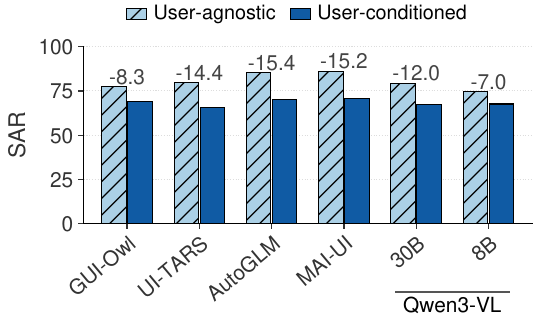}
            \caption{UA vs. UC contexts.}
            \label{fig:outcome_composition_a}
        \end{subfigure}
        \hfill
        \begin{subfigure}[t]{0.48\linewidth}
            \centering
            \includegraphics[width=\linewidth]
            {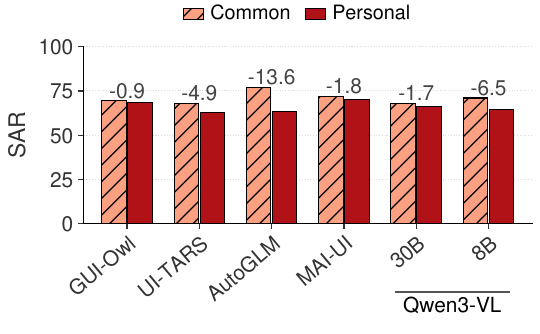}
            \caption{Common vs. personal targets.}
            \label{fig:outcome_composition_b}
        \end{subfigure}
        \vspace{-0.5em}
        \captionof{figure}{
            Subgoal achievement rate(SAR) across agents by (a) context type and (b) target type within user-conditioned contexts.
        }
        \label{fig:outcome_composition}
    \end{minipage}

\end{figure*}

\paragraph{Task success varies substantially across personas.}
We therefore examine how task outcomes vary across personas. Across the evaluated agents, as illustrated in Figure~\ref{fig:acrosspersona}, persona-level Task SR ranges from 43.8\% to 53.1\%, with a gap of up to 9.31 pp between the best- and worst-performing personas. This cross-persona variation occurs for both common targets with identical instructions and targets across users (6.94 pp) and personal targets drawn from each user’s own records (13.6 pp). Thus, the same task can produce different outcomes depending on the persona-conditioned interface in which it is executed.

\begin{figure*}[t]
    \centering

    % =========================
    % Figure 5
    % =========================
    \begin{minipage}[t]{0.32\textwidth}
        \centering

        \includegraphics[width=\linewidth]
        {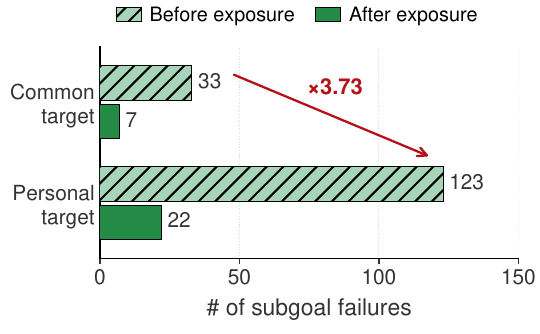}

        \captionof{figure}{
            Incorrect item selections among subgoal failures by target exposure and target type.
        }
        \label{fig:target_identity_actions}
    \end{minipage}
    \hfill
    \begin{minipage}[t]{0.66\textwidth}
    \centering

    \begin{subfigure}[t]{0.48\linewidth}
        \centering
        \includegraphics[
            width=\linewidth
        ]{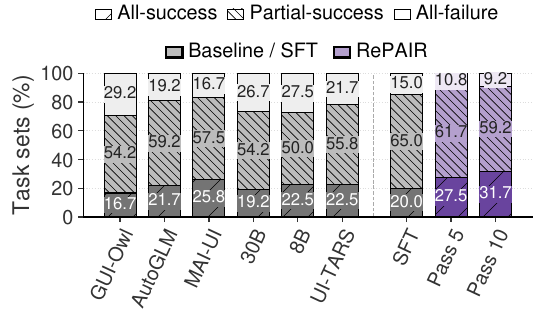}
        \caption{Task SR composition.}
        \label{fig:model1}
    \end{subfigure}
    \hfill
    \begin{subfigure}[t]{0.48\linewidth}
        \centering
        \includegraphics[
            width=\linewidth
        ]{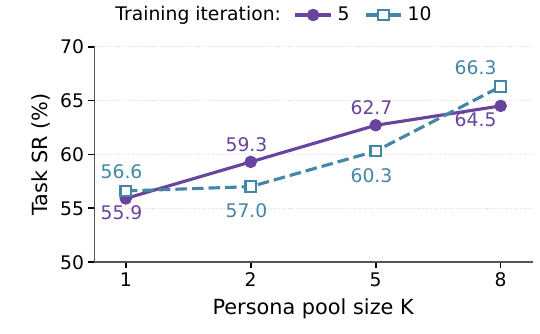}
        \caption{Task SR by user pool size.}
        \label{fig:model2}
    \end{subfigure}
    \vspace{-0.5em}
    \caption{\system{} results. (a) Task success consistency across users. (b) Task SR by training user pool size and training iteration.}
    \label{fig:model}
    \end{minipage}

\end{figure*}

% \begin{figure}[t]
%     \centering

%     \begin{subfigure}[t]{0.32\linewidth}
%         \centering
%         \includegraphics[
%             width=\linewidth
%         ]{figures/personalbench/method_cross_user_outcomes.pdf}
%         \caption{Target success composition.}
%         \label{fig:method_cross_user_outcomes}
%     \end{subfigure}
%     \hfill
%     \begin{subfigure}[t]{0.32\linewidth}
%         \centering
%         \includegraphics[
%             width=\linewidth
%         ]{figures/personalbench/persona_pool_ablation.pdf}
%         \caption{Performance by pool size at passes 5 and 10.}
%         \label{fig:nofprofile}
%     \end{subfigure}
%     \hfill

%     \caption{
%         \system{} results. (a) \system{} increases success across all users relative to baseline. (b) Performance improves with broader user-profile coverage and begins to saturate beyond XX profiles.
%     }
%     \label{fig:pgro_results}
% \end{figure}

\paragraph{Goal achievement drops in user-conditioned contexts, with difficulty for personal targets.}
Across all six complete agents, the SAR is consistently lower in user-conditioned than in user-agnostic contexts, with an average drop of 12.1 (SD 3.64) pp (Figure~\ref{fig:outcome_composition}(a)). Within user-conditioned contexts, personal targets yield lower SAR than common targets across all six agents (Figure~\ref{fig:outcome_composition}(b)). These results suggest task progress is more difficult in interface regions containing user-conditioned content, particularly when the target is resolved from semantically related items.

Beyond the aggregate SAR gap, the trajectory-level analysis reveals localized peaks in failure at specific subgoals (see Figure~\ref{fig:mixed18_ip_sar} in Appendix~\ref{app:subgoal_failure}). High failure rates frequently occur at subgoals conducted in user-conditioned contexts. This pattern appears across multiple applications and task types, suggesting that the difficulty is not confined to a single interface or operation. From a trajectory-level perspective, user-conditioned contexts can therefore act as major bottlenecks, disrupting task progress even when other stages are completed successfully.

\begin{wrapfigure}{R}{0.36\textwidth}
    \centering
    \small

    % (a)
    \includegraphics[
        width=\linewidth,
        trim=0bp 0bp 216bp 18bp,
        clip
    ]{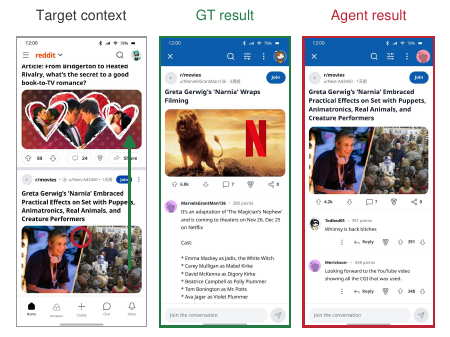}

    \par\vspace{1pt}
    \textbf{(a)} Before target exposure.

    \par\vspace{5pt}

    % (b)
    \includegraphics[
        width=\linewidth,
        trim=216bp 0bp 0bp 18bp,
        clip
    ]{figures/personalbench/finding3_failure_examples.pdf}

    \par\vspace{1pt}
    \textbf{(b)} After target exposure.

    \caption{
        Personal-target failures in user-conditioned contexts: the agent selects an incorrect item (a) before target exposure and (b) after the target becomes visible. Red: agent action; green: ground truth.
    }
    \label{fig:target_action_examples}

    \vspace{-3\intextsep}
\end{wrapfigure}

\paragraph {Personal targets increase competing-item selection errors.}
To better understand failures in user-conditioned contexts, we further analyze the error patterns that arise when agents resolve targets. For both common and personal targets, incorrect-item selection failures occur more frequently before target exposure, when further navigation is required, than after exposure, with personal targets exhibiting more failures overall (Figure~\ref{fig:target_identity_actions}).

% Subgoal failures involving selecting another list item instead of the intended target occur more often before target exposure than after exposure, for both common and personal targets (Figure~\ref{fig:target_identity_actions}). These failures are also more frequent for personal targets.

A closer inspection of these failures reveals two recurring patterns depending on whether the intended target has been exposed. Figure~\ref{fig:target_action_examples} illustrates representative personal target failures under explicit instructions. In the first case, the agent opens a similarly titled user-specific post before the intended target appears, rather than continuing to swipe. In the second case, the intended video is already visible, yet the agent selects a different video from the same personalized list. These examples indicate that agents may commit to a competing item before reaching the target, or fail to distinguish it from surrounding items even after it becomes visible. Notably, both failures occur despite the target being named explicitly, suggesting that exact target specification alone does not eliminate interference from competing user-specific context (Appendix~\ref{app:failure_protocol}).

\paragraph{\system{} improves cross-user reliability on unseen personas.}
% Our preceding analyses show cross-user variation in task success and lower subgoal achievement in user-conditioned contexts. 
We first adapt the base model to the benchmark tasks with supervised fine-tuning. This substantially improves SAR in user-agnostic contexts (+17.7 pp) but yields smaller gains in user-conditioned contexts (+4.24 pp) and no improvement in all-success, leaving difficulties in user-conditioned contexts and cross-user inconsistency unresolved.
We then apply \system{} to examine whether learning from cross-user differences improves reliability in user-conditioned contexts. \system{} improves user-conditioned SAR (+5.87 pp), the all-success rate (+7.50 pp), and overall Task SR (+9.42 pp) on unseen users. The gain in user-conditioned SAR indicates greater robustness to user-conditioned interface variation, while the increase in the all-success rate reflects more consistent task completion across users (see Figure~\ref{fig:model}(a)). 

The concurrent improvement in task success further shows that these gains translate to better end-to-end performance. To this end, we vary the training user pool size $n_{\mathrm{p}}$ and the number of training iterations. Increasing $n_{\mathrm{p}}$ from 1 to 8 improves overall Task SR after both 5 (+8.6 pp) and 10 (+9.7 pp) iterations (Figure~\ref{fig:model}(b)). However, extending training from 5 to 10 iterations decreases Task SR for $n_{\mathrm{p}}=2$ ($-2.3$ pp) and $n_{\mathrm{p}}=5$ ($-2.4$ pp), while improving it for $n_{\mathrm{p}}=8$ (+1.8 pp). These results suggest that the benefit of additional \system{} training depends on user-pool diversity, with repeated comparisons improving unseen-user performance only when training spans a broad set of users.

\section{Conclusion and Future Work}
% \ts{unclear -- edited}

We systematically study cross-user reliability in mobile GUI agents: whether the same request is completed consistently across user-conditioned application states. We introduce \bench{}, which constructs such states for controlled evaluation, and \system{}, which trains from cross-user differences in user-conditioned subgoal outcomes. Experiments reveal cross-user performance variation, lower subgoal achievement in user-conditioned contexts, especially for personal targets, and incorrect selections before and after target exposure; \system{} improves performance and consistency on unseen personas, providing initial evidence that cross-user differences can serve as a useful training signal.

These findings suggest several directions for future work. 1) Develop training methods that leverage cross-user differences in subgoal outcomes to improve reliability in user-conditioned contexts. 2) Strengthen target grounding through uncertainty-aware search, target verification, and recovery from incorrect-item selections before and after target exposure. 3) Determine the optimal balance between persona pool size and the number of training iterations to improve generalization to unseen users.

\FloatBarrier
\bibliography{iclr2027_conference}
\bibliographystyle{iclr2027_conference}

\clearpage
\appendix
% Compact four-part appendix. Earlier versions are retained in the backup.
\section{FingerTip-20K Analysis}
\label{app:fingertip_protocol}

\paragraph{Data.}
All six models are evaluated on the same 103,199 steps from the FingerTip-20K~\citep{yang2026fingertip}
training split: 12,016 episodes, 83 users and 102 app labels. We retain
supported actions with available screenshots from apps used by at least five
users. Each prediction uses the recorded screenshot, instruction and
human action history. Later screenshots follow the human demonstration.

\paragraph{Screen labels and segments.}
A text label is user-specific if its normalized form appears for only one
user within an app. A screen is user-conditioned (UC) if it contains such a
label. Numeric and time labels are included; icons and unknown labels are
excluded. Labels are not filtered by pixel visibility.
Consecutive screens form a segment until the Jaccard similarity of labels
outside scrollable regions falls below 0.475; two empty sets have similarity
one. This gives 60,383 segments: 51,761 UC and 8,622 without UC elements.

\paragraph{Scoring.}
An action must have the correct type. For clicks and long clicks, the pixel
distance from the reference position, divided by the image diagonal and
rounded to four decimals, must be at most 0.14. Scroll direction must match;
text is compared after removing whitespace and lowercasing. Other supported
actions are scored by type, and parsing failures are incorrect. A segment
succeeds only if all its actions are correct. All 60,383 segments are included.

\paragraph{Target exposure.}
The exposure analysis uses 29,640 segments with a labeled exit target.
A clickable element matches the target label if character-bigram Jaccard
similarity is at least 0.85, target-bigram coverage is at least 0.8, the shared
prefix has at least six characters, or the matching substring has at least
eight. This text-based measure does not establish visual attention.
Differences between screen groups are descriptive; tasks, apps and segment
lengths are not controlled.

\clearpage
\section{PAIR Implementation, Screens and Dataset Statistics}
\label{app:b_}
\subsection{User-conditioned application state generation}
\label{app:pair_implementation}
PAIR converts PersonaLens~\citep{zhao2025personalens} interests into app-specific queries. Content is
ranked by name matches (score 1) or cosine similarity using
\texttt{paraphrase-multilingual-MiniLM-L12-v2}~\citep{reimers-2019-sentence-bert} with mean pooling,
L2 normalization and a 96-token limit. Duplicate IDs/names and reserved
Common items are removed before selection. Common items are then added,
and lists are shuffled using seed-0 hashes. Most lists contain 20 items.
App-specific generators also create transactions, chats and meetings.
The records keep their IDs and media links and are rendered by MobileGym~\citep{wu2026mobilegym}.

\begin{figure}[H]
\centering
\includegraphics[width=0.93\linewidth]{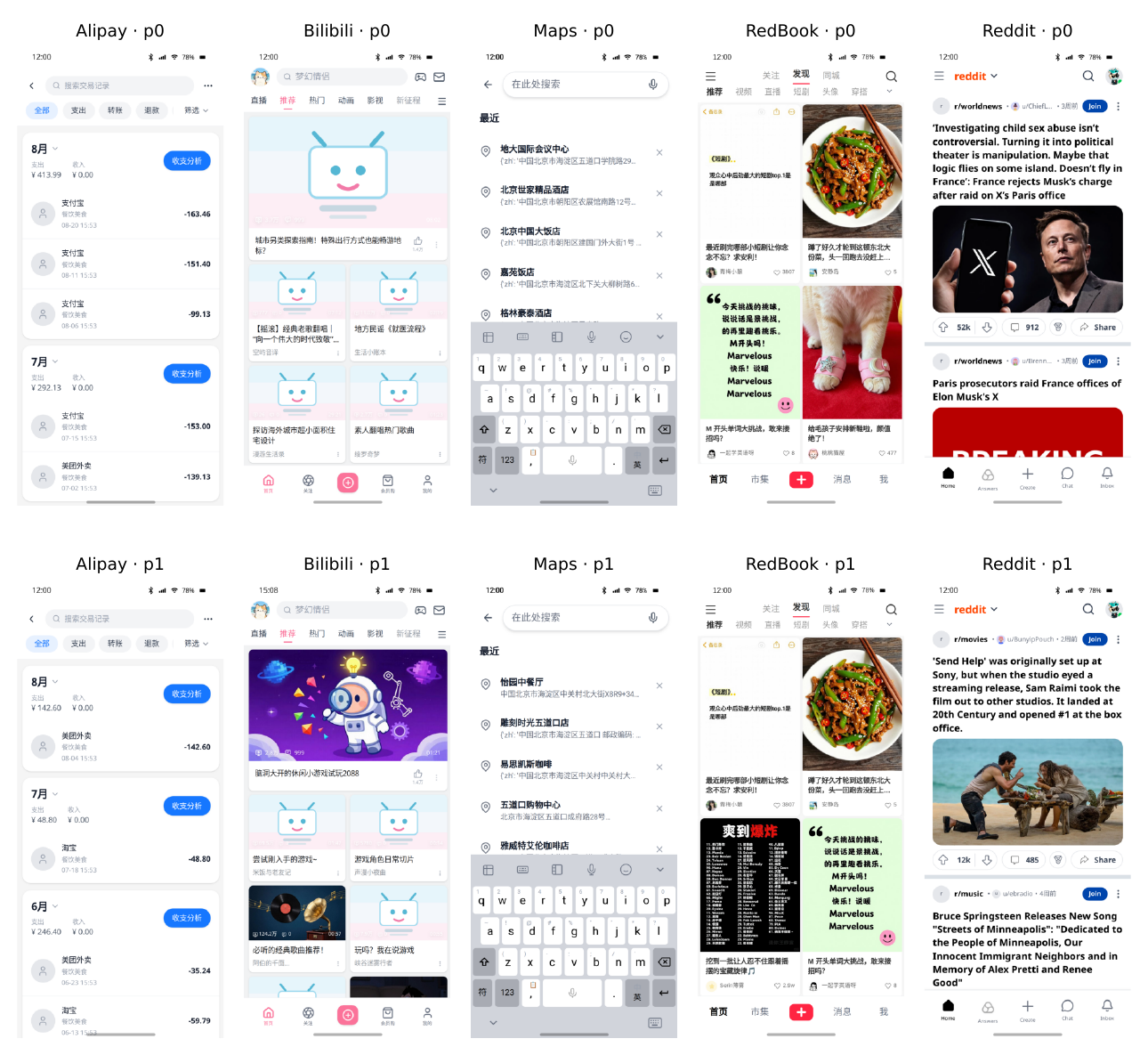}
\caption{Generated screens for p0 (top) and p1 (bottom). Each column shows
the same app and screen location for both personas.}
\label{fig:pair_app_screens}
\end{figure}

\clearpage
\begin{figure}[H]
\centering
\includegraphics[width=0.93\linewidth]{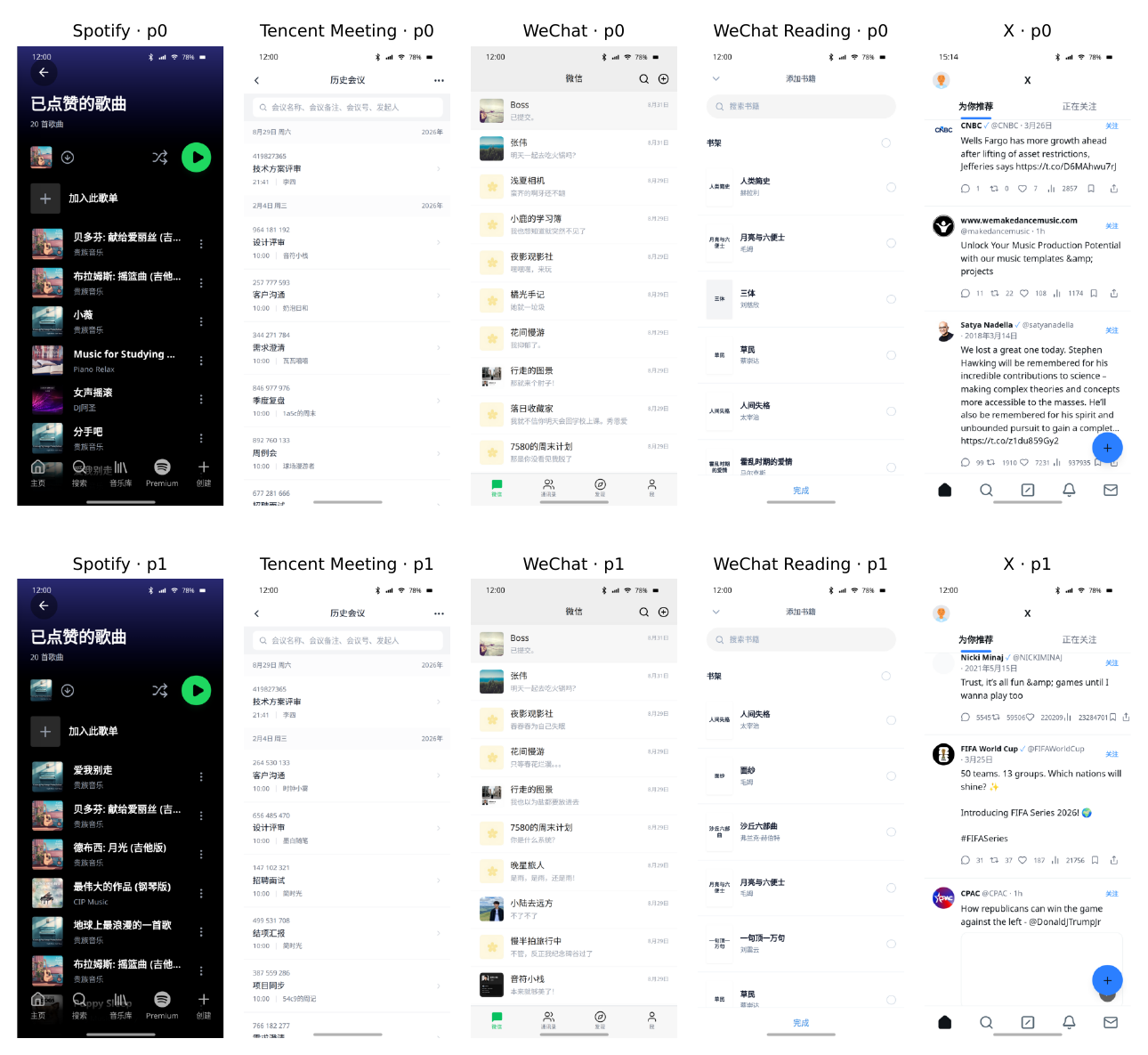}
\caption{Persona comparison for the remaining five apps: p0 (top), p1 (bottom).}
\label{fig:pair_app_screens_remaining}
\end{figure}

% \paragraph{Dataset size.}
% PAIR has 20 personas, ten apps and 30 task templates (three per app).
% Common/Personal targets and Explicit/Implicit instructions give
% $20\times30\times2\times2=2{,}400$ task instances.
% Train, validation and test use p10 to p17, p18 to p19 and p0 to p9, respectively.
% Personas are disjoint; apps, templates and content sources are shared.

% \begin{table}[H]\centering\small
% \caption{PAIR data counts. Each of the four conditions has equal size.
% Apps and templates are shared across splits.}
% \label{tab:pair_dataset_statistics}
% \begin{tabular}{lrrrr}
% \toprule
% Statistic & Train & Validation & Test & All \\
% \midrule
% Personas & 8 & 2 & 10 & 20 \\
% Applications & 10 & 10 & 10 & 10 \\
% Task templates & 30 & 30 & 30 & 30 \\
% Instances per condition & 240 & 60 & 300 & 600 \\
% Total task instances & 960 & 240 & 1,200 & 2,400 \\
% \bottomrule
% \end{tabular}
% \end{table}

\clearpage
% \subsection{Apps and Task Templates}
% \label{app:task_catalog}
\subsection{Task generation}
\label{app:generated_data}

Common targets are shared across personas; Personal targets come from each
persona's data. Explicit and Implicit instructions refer to the same target.
All four conditions use the same initial state for a template/persona pair,
reset before each episode. The instruction writer (\texttt{openai/gpt-oss-120b}~\citep{openai2025gptoss120bgptoss20bmodel})
and reviewer use the candidate records. Agents receive only screenshots,
instructions and action history.

All 1,200 test instances are evaluated against their designated targets,
including cases with unresolved review flags. Automatic review does not
guarantee a unique interpretation of every Implicit instruction.

\begin{table}[H]
    \centering
    \small
    \caption{Apps and target screens. RedBook is also called Xiaohongshu.}
    \label{tab:apps_surfaces}
    \begin{tabularx}{\linewidth}{@{}lX@{}}
        \toprule
        Application & Target UI contexts \\
        \midrule
        Alipay & Billing history; message list \\
        Bilibili & Subscriptions; home recommendations; liked videos \\
        Maps & Recent searches \\
        RedBook & Saved notes; liked notes; discover feed \\
        Reddit & Home feed \\
        Spotify & Liked songs; recently played tracks; home playlists \\
        Tencent Meeting & Meeting history; scheduled meetings \\
        WeChat & Chat list; Moments \\
        WeChat Reading & Home reading list; Add Favorite Book; recommendations \\
        X & For You feed \\
        \bottomrule
    \end{tabularx}
\end{table}

\begingroup
\small
\setlength{\tabcolsep}{4pt}
\begin{longtable}{@{}p{2.6cm}p{2.1cm}p{1.7cm}>{\raggedright\arraybackslash}p{6.2cm}@{}}
\caption{All 30 task templates. Each operation applies to the designated
Common/Personal target under Explicit/Implicit instructions.}
\label{tab:task_families}\\
\toprule
\textbf{Application} & \textbf{Task ID} & \textbf{Target} & \textbf{Task Template} \\
\midrule
\endfirsthead

\toprule
\textbf{Application} & \textbf{Task ID} & \textbf{Target} & \textbf{Task Template} \\
\midrule
\endhead

\bottomrule
\endfoot

Alipay
& \texttt{GG-ALI-01}
& Transaction
& Open the transaction in billing history. \\

Alipay
& \texttt{GG-ALI-02}
& Conversation
& Open the conversation in the message list. \\

Alipay
& \texttt{GG-ALI-03}
& Conversation
& Open the conversation; reply exactly ``Understood''. \\

\midrule
Bilibili
& \texttt{GG-BIL-01}
& Video
& Open the subscription video and like it. \\

Bilibili
& \texttt{GG-BIL-02}
& Video
& Open the home recommendation video. \\

Bilibili
& \texttt{GG-BIL-03}
& Video
& Open the video in Liked Videos. \\

\midrule
Maps
& \texttt{GG-MAP-01}
& Place
& Open place details from recent searches. \\

Maps
& \texttt{GG-MAP-02}
& Place
& Get directions to the place in recent searches. \\

Maps
& \texttt{GG-MAP-03}
& Place
& Open place details from recent searches. \\

\midrule
RedBook
& \texttt{GG-RBK-01}
& Note
& Open the saved note. \\

RedBook
& \texttt{GG-RBK-02}
& Note
& Open the liked note. \\

RedBook
& \texttt{GG-RBK-03}
& Note
& Open the discover-feed note and save it. \\

\midrule
Reddit
& \texttt{GG-RDT-01}
& Post
& Open the home-feed post. \\

Reddit
& \texttt{GG-RDT-02}
& Post
& Open the post; comment exactly ``Thank you''. \\

Reddit
& \texttt{GG-RDT-03}
& Post
& Open the home-feed post. \\

\midrule
Spotify
& \texttt{GG-SPT-01}
& Song
& Add the liked song to a new ``Rainy Night'' playlist. \\

Spotify
& \texttt{GG-SPT-02}
& Song
& Play the recently played song. \\

Spotify
& \texttt{GG-SPT-03}
& Playlist
& Open the home-screen playlist. \\

\midrule
Tencent Meeting
& \texttt{GG-MTG-01}
& Meeting
& Open the meeting in history. \\

Tencent Meeting
& \texttt{GG-MTG-02}
& Meeting
& Open the scheduled meeting. \\

Tencent Meeting
& \texttt{GG-MTG-03}
& Meeting
& Open the meeting in history. \\

\midrule
WeChat
& \texttt{GG-WCT-01}
& Chat
& Open the chat. \\

WeChat
& \texttt{GG-WCT-02}
& Chat
& Open the chat; reply exactly ``Understood''. \\

WeChat
& \texttt{GG-WCT-03}
& Author
& Open the author's profile from Moments. \\

\midrule
WeChat Reading
& \texttt{GG-WRE-01}
& Book
& Open the book in the home reading list. \\

WeChat Reading
& \texttt{GG-WRE-02}
& Book
& Add the book using Add Favorite Book. \\

WeChat Reading
& \texttt{GG-WRE-03}
& Book
& Open the recommended book. \\

\midrule
X
& \texttt{GG-X-01}
& Post
& Open the For You post. \\

X
& \texttt{GG-X-02}
& Post
& Open the For You post. \\

X
& \texttt{GG-X-03}
& Post
& Open the For You post. \\

\end{longtable}
\endgroup
% \clearpage
\subsection{Reference Subgoals by Task}
\label{app:task_subgoals}

Table~\ref{tab:task_subgoals} expands the task templates in Table~\ref{tab:task_families} into their ordered reference subgoals. Each subgoal specifies an objective associated with a UI context and may involve multiple actions, including scrolling, typing, and clicking. The sequence is shared across Common/Personal targets and Explicit/Implicit instructions; the designated target is determined by the task instance. These descriptions specify the reference objectives; the RePAIR training reward is defined separately in Appendix~\ref{app:reward_implementation}.

\begingroup
\small
\setlength{\tabcolsep}{4pt}
\renewcommand{\arraystretch}{1.15}
\begin{longtable}{@{}p{2.2cm}>{\raggedright\arraybackslash}p{\dimexpr\linewidth-2.2cm-2\tabcolsep\relax}@{}}
\caption{The 67 reference subgoal positions across all 30 task templates included in our analysis. The optional library-to-history navigation step in GG-SPT-02 is excluded, and the remaining positions are numbered consecutively within each task.}
\label{tab:task_subgoals}\\
\toprule
\textbf{Task ID} & \textbf{Ordered subgoals} \\
\midrule
\endfirsthead
\toprule
\textbf{Task ID} & \textbf{Ordered subgoals (continued)} \\
\midrule
\endhead
\midrule
\multicolumn{2}{r}{\emph{Continued on next page}} \\
\endfoot
\bottomrule
\endlastfoot
\texttt{GG-ALI-01} & (1) Open My from the home screen; (2) Open billing history; (3) Open the designated transaction details. \\
\texttt{GG-ALI-02} & (1) Open the message list from the home screen; (2) Open the designated conversation. \\
\texttt{GG-ALI-03} & (1) Open the message list from the home screen; (2) Open the designated conversation; (3) Enter and send ``Understood''. \\
\midrule
\texttt{GG-BIL-01} & (1) Open Following from the recommendation home screen; (2) Switch to the video tab; (3) Open the designated video; (4) Like the designated video. \\
\texttt{GG-BIL-02} & (1) Open the designated video from the home recommendation feed. \\
\texttt{GG-BIL-03} & (1) Open Me from the recommendation home screen; (2) Open the personal space; (3) Open the liked-video list; (4) Open the designated video. \\
\midrule
\texttt{GG-MAP-01} & (1) Open search from the map home screen; (2) Open the designated place details from recent searches. \\
\texttt{GG-MAP-02} & (1) Open search from the map home screen; (2) Select the designated place from recent searches; (3) Open directions to the designated place. \\
\texttt{GG-MAP-03} & (1) Open search from the map home screen; (2) Open the designated place details from recent searches. \\
\midrule
\texttt{GG-RBK-01} & (1) Open Me from the discovery home screen; (2) Open the saved-notes tab; (3) Open the designated note. \\
\texttt{GG-RBK-02} & (1) Open Me from the discovery home screen; (2) Open the liked-notes tab; (3) Open the designated note. \\
\texttt{GG-RBK-03} & (1) Open the designated note from the discovery feed; (2) Save the designated note. \\
\midrule
\texttt{GG-RDT-01} & (1) Open the designated post from the home feed. \\
\texttt{GG-RDT-02} & (1) Open the designated post from the home feed; (2) Enter and submit the comment ``Thank you''. \\
\texttt{GG-RDT-03} & (1) Open the designated post from the home feed. \\
\midrule
\texttt{GG-SPT-01} & (1) Open the library from the home screen; (2) Open Liked Songs; (3) Open the menu for the designated track; (4) Select Add to playlist; (5) Select the option to create a new playlist; (6) Create the playlist named ``Rainy Night'' and complete the track addition. \\
\texttt{GG-SPT-02} & (1) Open the library from the home screen; (2) Expand the recent-play group; (3) Play the designated track within the expanded group. \\
\texttt{GG-SPT-03} & (1) Open the designated playlist from the home screen. \\
\midrule
\texttt{GG-MTG-01} & (1) Open meeting history from the home screen; (2) Open the designated meeting details. \\
\texttt{GG-MTG-02} & (1) Open the designated scheduled meeting details from the home screen. \\
\texttt{GG-MTG-03} & (1) Open meeting history from the home screen; (2) Open the designated meeting details. \\
\midrule
\texttt{GG-WCT-01} & (1) Open the designated conversation from the chat list. \\
\texttt{GG-WCT-02} & (1) Open the designated conversation from the chat list; (2) Enter and send ``Understood''. \\
\texttt{GG-WCT-03} & (1) Open Discover from the chat list; (2) Open Moments; (3) Open the designated author's profile. \\
\midrule
\texttt{GG-WRE-01} & (1) Open the designated book from the home reading list. \\
\texttt{GG-WRE-02} & (1) Open Me from the reading home screen; (2) Open the book-list screen; (3) Open the list editor; (4) Open Add Favorite Book; (5) Select the designated book and confirm its addition. \\
\texttt{GG-WRE-03} & (1) Open the designated book from the home recommendations. \\
\midrule
\texttt{GG-X-01} & (1) Open the designated post from the For You feed. \\
\texttt{GG-X-02} & (1) Open the designated post from the For You feed. \\
\texttt{GG-X-03} & (1) Open the designated post from the For You feed. \\
\end{longtable}
\endgroup

% \clearpage
\section{Training Rewards and Algorithm}
\label{app:reward_implementation}

\paragraph{Reward calculation.}
We compute a binary reward for each subgoal from the recorded trajectory. A subgoal first requires evidence that its entry screen was observed before an action; appearing only in the final screen is insufficient. For a nonterminal subgoal, reward one additionally requires either a recorded route corresponding to a later subgoal in the reference path or a click on an expected item or control. For the terminal subgoal, defined as the last subgoal in the reference path, reward one instead requires the task judge to report success. Reward labels and matching criteria are used only for scoring and are not provided to the agent.

For rollout $i$ and subgoal $k$, let $E_{ik}$ indicate the required entry observation, $L_{ik}$ a matching later-reference route in any observation or the final screen, and $C_{ik}$ an expected click anywhere in the log. Let $J_i$ be the binary task-judge outcome. The reward is
\begin{equation}
 r_{ik}=\begin{cases}
 0, & E_{ik}=0,\\
 J_i, & E_{ik}=1\text{ and }k\text{ is terminal},\\
 1, & E_{ik}=1,\ k\text{ nonterminal, and }(L_{ik}\lor C_{ik}),\\
 0, & \text{otherwise}.
 \end{cases}
\label{eq:reward_implementation}
\end{equation}

\paragraph{Matching rules.}
Route matching compares URL paths after removing query parameters; app identity and screen state are not used. Expected items and controls are specified by the task template. A target-item click matches when its recorded IDs include the designated target ID or identity. Other controls match by trigger ID or by a substring match in either direction between whitespace-normalized labels. The scorer searches the entire log without enforcing visit order. Here, ``later'' refers to the order of subgoals in the reference path, not the order of observations in the rollout.

\paragraph{Policy update.}
Each iteration collects 124 rollouts with a fixed policy and performs one LoRA update per four-rollout group. Rollout-policy and SFT-reference token log probabilities are cached as specified in the pseudocode below.

\begin{table}[H]
\centering\small
\caption{RePAIR training pseudocode ($K=8$; $N\in\{5,10\}$ training iterations). $\theta$ denotes trainable LoRA parameters; $h_{it}$ is the recorded token prefix and $y_{it}$ a response token. $T_i$ counts response tokens over the entire trajectory, across all actions.}
\label{tab:repair_training_pseudocode}
\begin{tabularx}{\linewidth}{@{}r@{\quad}X@{}}
\toprule
1 & $\theta \leftarrow \mathrm{SFT}(\theta_{\mathrm{base}},\mathcal D_{\mathrm{Explicit}})$ for one epoch;
freeze $\theta_{\mathrm{ref}}\leftarrow\theta$. \\
2 & Build fixed persona pools and a repeating five-iteration pair schedule. \\
3 & \textbf{for} training iteration $j=1,\ldots,N$ \textbf{do} \\
4 & \quad Fix rollout policy $\theta_{\mathrm{old}}\leftarrow\theta$ for this iteration. \\
5 & \quad \textbf{for each} of the 31 task--condition pairs $q$ \textbf{do} \\
6 & \qquad Select the scheduled pair; collect two full rollouts per slot with vLLM. \\
7 & \qquad Cache the four trajectories and their rollout-policy token log probabilities. \\
8 & \quad Cache SFT-reference token log probabilities; reset AdamW and the 31-update cosine schedule. \\
9 & \quad \textbf{for each} cached group in fixed task order \textbf{do} \\
10 & \qquad Compute subgoal rewards $r_{ik}$ using Section~\ref{app:reward_implementation}. \\
11 & \qquad $a_{ik}\leftarrow(r_{ik}-\mu_k)/\sqrt{\sigma_k^2+10^{-8}}$,
using the mean and population variance over $i=1,\ldots,4$. \\
12 & \qquad $A_i\leftarrow |\mathcal U_q|^{-1}\sum_{k\in\mathcal U_q}a_{ik}$;
$\mathcal U_q$ is the UC-subgoal set (use zero if empty). \\
13 & \qquad Recompute current-policy token log probabilities; reuse cached SFT-reference scores. \\
14 & \qquad $\rho_{it}\leftarrow\pi_\theta(y_{it}\mid h_{it})/\pi_{\mathrm{old}}(y_{it}\mid h_{it})$;
$d_{it}\leftarrow\log\pi_{\mathrm{ref}}(y_{it}\mid h_{it})-\log\pi_\theta(y_{it}\mid h_{it})$. \\
15 & \qquad $\ell_{it}\leftarrow-\min\{\rho_{it}A_i,\mathrm{clip}(\rho_{it},0.8,1.2)A_i\}
+0.01(e^{d_{it}}-d_{it}-1)$. \\
16 & \qquad Take one AdamW step on $\frac14\sum_{i=1}^4\frac1{T_i}\sum_{t=1}^{T_i}\ell_{it}$;
clip gradient norm to 1.0 and advance the scheduler. \\
17 & \quad Save the iteration checkpoint. \\
18 & \textbf{return} the iteration-$N$ adapter ($124N$ rollouts; $31N$ optimizer updates). \\
\bottomrule
\end{tabularx}
\end{table}

\clearpage
\section{Experimental and Training Setup}
\label{app:experiment_details}

\paragraph{Evaluation.}
We evaluate Qwen3-VL-8B, Qwen3-VL-30B-A3B, GUI-Owl-1.5-8B-Think, MAI-UI-8B, AutoGLM-Phone-9B-Multilingual and UI-TARS-1.5-7B on the same 1,200 test instances. Decoding uses temperature 0, top-p 1 and seed 0, with each model's action format. The action limit is 20 for 840 instances and 30 for 360. Output limits are 1,024 tokens for Qwen/SFT/RePAIR, 2,048 for MAI-UI and 3,000 for AutoGLM; GUI-Owl/UI-TARS limits are not specified in the saved request configs. Task SR uses all instances; all-success uses 120 ten-persona groups; SAR uses reached subgoals at the 67 reference positions.

\paragraph{SFT.}
Qwen3-VL-8B-Instruct is fine-tuned on successful Explicit-instruction oracle demonstrations across all 30 task templates. Of the 480 Explicit task instances from the eight training personas, three failed oracle executions are excluded, leaving 477 demonstrations containing 1,876 actions. Validation uses 119 successful demonstrations containing 469 actions, after excluding one failed execution from 120 instances. All four excluded executions are from X and failed to reach the designated target route.

Implicit instructions, designated targets, and initial states are available, but separate Implicit oracle demonstrations were not collected. SFT therefore uses only Explicit demonstrations. LoRA~\citep{hu2021lora} updates the language model's q/k/v/o attention projections; the vision encoder is frozen. One epoch gives 235 updates using assistant-response loss. Training uses bf16, SDPA and gradient checkpointing on an RTX 6000 Ada (PyTorch 2.13.0+cu130, Transformers 4.57.6, PEFT 0.18.1).

\begin{table}[H]\centering\small
\caption{Core optimization settings. RePAIR inherits the SFT LoRA adapter.}
\label{tab:appendix_training_settings}
\begin{tabular}{lll}
\toprule
Setting & SFT & RePAIR \\
\midrule
LoRA rank / alpha & 16 / 32 & 16 / 32 \\
Dropout & 0.05 & Disabled during updates \\
Optimizer & AdamW~\citep{loshchilov2018decoupled} & AdamW \\
Learning rate & $10^{-4}$ & $10^{-5}$ \\
Weight decay / norm clipping & 0 / 1.0 & 0 / 1.0 \\
Scheduler / warmup & Cosine / 3\% & Iteration-wise cosine / 3\% \\
Update unit & 8 action samples & 4 full-task rollouts \\
Policy ratio clip & --- & 0.2 \\
Reference KL coefficient & --- & 0.01 (fixed SFT) \\
\bottomrule
\end{tabular}
\end{table}

\paragraph{RePAIR task selection.}
A task specification is a combination of task template, target type (Common or Personal), and instruction type (Explicit or Implicit). The 30 templates yield 120 specifications, each evaluated across ten personas, p0 to p9. For each model, a specification is classified as mixed when its final-goal outcomes include both successes and failures across these personas.

We select specifications classified as mixed by all five baselines in the frozen September 16, 2026 analysis: Qwen3-VL-8B, Qwen3-VL-30B-A3B, GUI-Owl, MAI-UI, and AutoGLM. This yields 31 specifications spanning 18 templates, comprising 12 Common and 19 Personal specifications, or 8 Explicit and 23 Implicit specifications.

Selection uses the historical AutoGLM results preceding its greedy rerun and excludes UI-TARS. The selected list is held fixed across subsequent training experiments rather than recomputed from updated baseline results.

\paragraph{RePAIR training.}
The 31 selected specifications are instantiated for the eight training personas p10 to p17, yielding 248 distinct task instances. Each training iteration collects four full-task rollouts per specification, totaling 124 rollouts and 31 optimizer updates. Both Explicit and Implicit tasks are included: the policy generates trajectories that are scored using recorded routes, actions, and the task judge, without requiring oracle demonstrations.

Seeded selection forms fixed pools of $K\in\{1,2,5,8\}$ personas. K1 uses one persona, K2 a fixed pair, and K5/K8 cycle pairs; four Reddit task--condition pairs allow same-persona sampling to keep subgoal labels consistent. Group compatibility requires matching ordered subgoal IDs and UA/UC labels across the full reference path. For each scheduled pair, we collect two rollouts per persona slot, resetting the corresponding task instance to its stored initial state before every rollout. Only UC-subgoal rewards are used to compute the trajectory advantage. Rollouts use temperature 0.8, top-p 1, top-k $-1$, min-p 0 and a 1,024-token output limit. Five/ten iterations give 620/1,240 rollouts and 155/310 updates. Iterations 6 to 10 repeat the persona schedule with new rollouts, continuing from iteration 5. The main result uses K8 at iteration 5; validation is not used to select it. SFT uses seed 0; RePAIR uses seed $j-1$ at iteration $j$ with derived update seeds.

\paragraph{Selection and review scope.}
Training rollouts use personas p10 to p17, disjoint from the evaluation personas p0 to p9. However, task specifications were selected using earlier baseline outcomes on p0 to p9. Thus, evaluation personas are held out from policy optimization, but their baseline results informed training-task selection. We interpret the RePAIR study as an exploratory intervention informed by the preceding analysis.

The p10 to p19 inputs have not completed semantic and browser review. Successful oracle execution and subsequent model rollouts do not establish completion of that review.

\paragraph{vLLM.}
We use vLLM~\citep{kwon2023efficient} for inference, with server seed 0 and GPU memory utilization 0.9. Qwen baselines use an 8,192-token batch budget and image bounds of 3,136 to 12,845,056 pixels; AutoGLM uses at most 5,000,000 pixels. K8 rollout logs for iterations 1 to 5 record vLLM 0.19.0, bf16, prefix caching and chunked prefill. These settings are not verified for all earlier FingerTip or recovery runs.

\begin{table}[H]\centering\small
\caption{vLLM settings from evaluation configs and training logs. TP: tensor parallelism; Context: token limit; Sequences/Images: per-server and per-request limits, respectively.}
\label{tab:appendix_vllm}
\begin{tabular}{lrrrr}
\toprule
Model / use & TP & Context & Sequences & Images \\
\midrule
Qwen3-VL-8B baseline & 1 & 16,384 & 8 & 1 \\
Qwen3-VL-30B-A3B & 2 & 16,384 & 8 & 1 \\
GUI-Owl & 1 & 32,768 & 8 & 5 \\
MAI-UI & 1 & 32,768 & 8 & 4 \\
UI-TARS & 1 & 32,768 & 8 & 6 \\
AutoGLM (PAIR / FingerTip) & 1 & 25,480 & 16 / 32 & 10 \\
SFT/RePAIR evaluation & 1 & 32,768 & 8 & 2 \\
RePAIR rollout collection & 1 & 32,768 & 8 & 2 \\
\bottomrule
\end{tabular}
\end{table}
% \clearpage
\section{Subgoal Failure Patterns Across Task Trajectories}
\label{app:subgoal_failure}

To examine where task progress breaks down, we report subgoal-level
failure rates along the reference trajectory for all 30 task templates.
For each task, subgoals are ordered according to their position in the
reference execution path. We compute the failure rate at each subgoal as
$100-\mathrm{SAR}$ after the agent reaches the corresponding subgoal
context. Thus, failure rates at later positions are conditional on the
agent having progressed far enough to reach those contexts.

\begin{figure}[!ht]
    \centering
    \includegraphics[
        width=0.72\linewidth,
        height=0.72\textheight,
        keepaspectratio
    ]{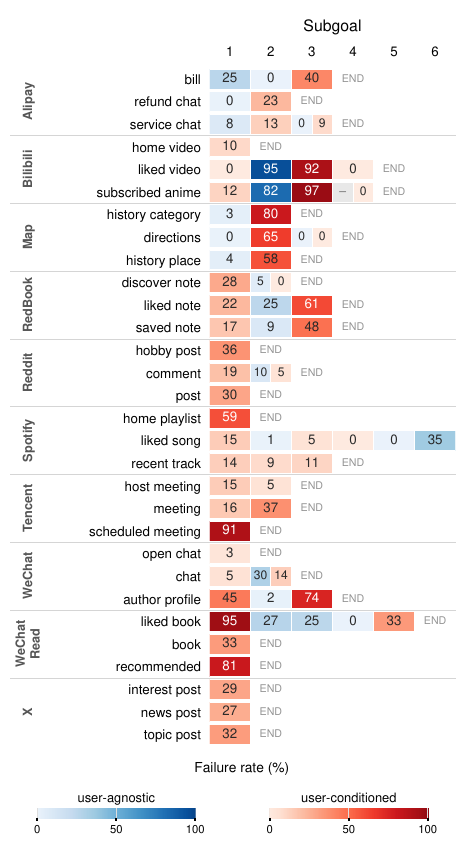}

    \caption{ 
        Subgoal failure rates across task trajectories.
        Each cell reports the failure rate ($100-\mathrm{SAR}$, \%)
        after entering the corresponding subgoal context, pooled across
        six complete agents. Blue and red denote user-agnostic and
        user-conditioned contexts, respectively; darker cells indicate
        higher failure rates. A dash denotes no entry, and END marks
        task termination. Later-position rates are conditional on
        reaching the corresponding context.
    }
    \label{fig:mixed18_ip_sar}
\end{figure}

Figure~\ref{fig:mixed18_ip_sar} shows that failures are not distributed
uniformly across task trajectories. Instead, many tasks exhibit localized
failure peaks at particular subgoals. Several pronounced failures occur
in user-conditioned contexts, including target-related stages for
Bilibili, Maps, Tencent Meeting, WeChat, and WeChat Reading. For example,
the failure rate reaches 100\% at the third subgoal of the Bilibili
\textit{subscribed anime} task, 90\% at the first subgoal of
\textit{scheduled meeting}, and 94\% at the first subgoal of
\textit{liked book}. These patterns indicate that particular stages
involving user-specific application content can form substantial
bottlenecks within otherwise multi-step tasks.

At the same time, high failure rates are not exclusive to
user-conditioned contexts. Some user-agnostic stages also show large
failure rates, such as the second subgoal of the Bilibili
\textit{liked video} and \textit{subscribed anime} tasks. Thus, the
trajectory-level analysis does not imply that all failures originate
from personalization. Rather, it distinguishes general GUI interaction
difficulties from localized failures that arise at user-conditioned
stages.

Overall, this analysis complements the aggregate SAR comparison by
showing where the observed performance differences emerge within
individual task trajectories. The results suggest that difficulty is
often concentrated at specific interaction stages rather than being
uniformly distributed throughout an entire task.
% \clearpage

\section{Complete PAIR Failure Trajectories}
\label{app:failure_protocol}

\label{app:finding3_protocol}

The two failures from the main paper use Personal targets and Explicit
instructions. Each row shows all three recorded steps, including termination.
Screenshots precede the labeled actions; no post-termination screenshot is available.

\begin{figure}[H]
\centering
\includegraphics[width=\linewidth]{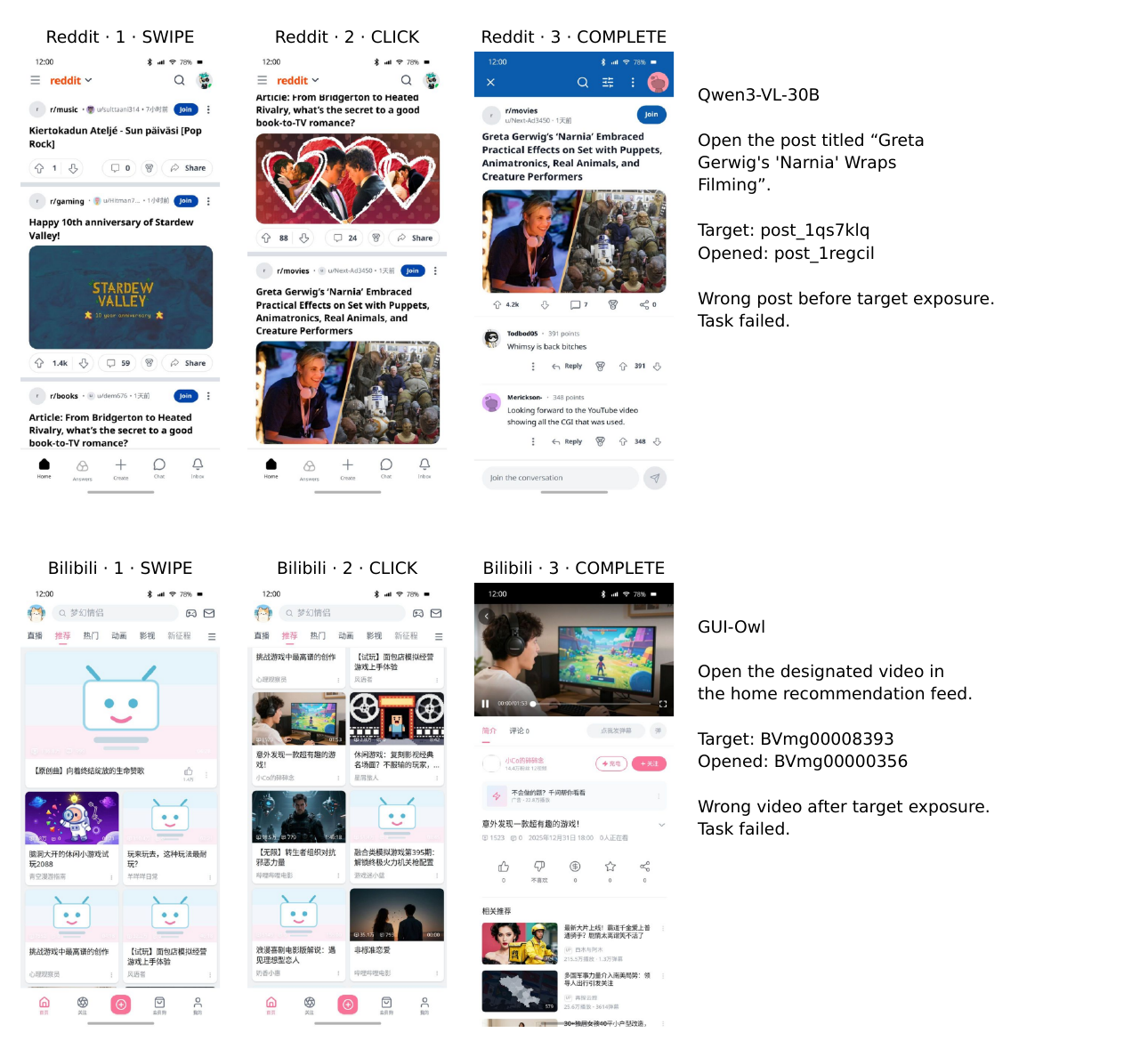}
\caption{Reddit (top, Qwen3-VL-30B) selects the wrong post before target
exposure. Bilibili (bottom, GUI-Owl) selects the wrong video after target
exposure. Both fail the task's success check; target and selected IDs are
shown beside the trajectories.}
\label{fig:pair_full_failure_trajectories}
\end{figure}

The selected item is confirmed by the next screen's route. These examples
illustrate the two error types, not their overall frequency.

\end{document}